\documentclass[12pt]{article}
\usepackage[utf8]{inputenc}
\usepackage{bm}
\usepackage{indentfirst} 
\usepackage{color}
\usepackage{ulem}
\usepackage{amsmath} 
\usepackage{caption}
\usepackage{subcaption}
\usepackage{float}
\usepackage{booktabs}
\usepackage{amssymb}
\usepackage{algorithm}
\usepackage{algorithmic}
\usepackage{graphicx}
\usepackage{tabularx}
\usepackage{multirow}
\usepackage[font=scriptsize, labelfont=bf]{caption}
\usepackage[numbers, square, comma, sort&compress]{natbib}
\usepackage{geometry}
\def\d{{\bm{d}}}
\def\z{{\bm{z}}}
\def\x{{\bm{x}}}

\renewcommand{\cite}[1]{\textsuperscript{\citep{#1}}} 
\title{Accelerated Bayesian Optimal Experimental Design via Conditional Density Estimation and Informative Data}

\author{ Miao Huang$^a$\quad Hongqiao Wang$^{a,b}$ \quad Kunyu Wu$^{a,*}$  \ \\
{\it $^a$School of Mathematics and Statistics} \\
{\it Central South University} \\
{\it Changsha 410083, People’s Republic of China}\\
[2mm]
{\it $^b$Institute of Mathematics} \\
{\it Henan Academy of Sciences} \\
{\it Zhengzhou 450046, People’s Republic of China}\\
[2mm]
{$^*$ Corresponding author: Kunyu Wu}\\
{\it E-mail: 242101036@csu.edu.cn}}

\date{}

\begin{document}
\maketitle
\maketitle {\flushleft\large\bf Abstract: }

The Design of Experiments (DOEs) is a fundamental scientific methodology that provides researchers with systematic principles and techniques to enhance the validity, reliability, and efficiency of experimental outcomes. 
In this study, we explore optimal experimental design within a Bayesian framework, utilizing Bayes' theorem to reformulate the utility expectation—originally expressed as a nested double integral—into an independent double integral form, significantly improving numerical efficiency.
To further accelerate the computation of the proposed utility expectation, conditional density estimation is employed to approximate the ratio of two Gaussian random fields, while covariance serves as a selection criterion to identify informative data-set during model fitting and integral evaluation.
In scenarios characterized by low simulation efficiency and high costs of raw data acquisition, key challenges such as surrogate modeling, failure probability estimation, and parameter inference are systematically restructured within the Bayesian experimental design framework. The effectiveness of the proposed methodology is validated through both theoretical analysis and practical applications, demonstrating its potential for enhancing experimental efficiency and decision-making under uncertainty.


\section{Introduction} 

Bayesian Optimal Experimental Design (BOED) is an advanced statistical framework that combines Bayesian inference with decision theory to systematically optimize the selection of experimental conditions. The core objective of BOED is to identify the most informative experiments by maximizing a predefined utility function, which quantifies the expected benefit or information gain from a given design. By incorporating prior knowledge and updating beliefs through Bayes’ theorem, BOED efficiently accounts for uncertainty in both model parameters and observations, thereby guiding experimental decisions in a principled manner.

This framework is commonly employed to determine critical experimental parameters such as sample size, sampling locations, timing, and frequency. Its strength lies in its ability to integrate prior information with new experimental data, enabling more accurate parameter estimation while simultaneously reducing experimental cost and effort. Owing to its flexibility and robustness, BOED is particularly well-suited for complex systems where uncertainty is inherent, and it has proven effective in yielding more informative and reliable experimental outcomes across a broad range of scientific and engineering applications.

The concept of Design of Experiments (DOE) was first introduced by the statistician Ronald Fisher in agricultural trials during the 1920s \citep{fisher1922mathematical}, and has since seen rapid and comprehensive development across various scientific disciplines. Initially, DOE methods were grounded in the classical experimental design framework, which included techniques such as orthogonal design and sequential design. 
Optimality criteria under this framework were typically based on the Fisher information matrix \citep{atkinson1996usefulness, pukelsheim2006optimal, fedorov2013theory}, leading to the development of A-optimal \citep{sim2005global} and D-optimal\citep{de1995d} design methods, among others. 

As DOE gained prominence, researchers recognized the limitations of classical experimental design, particularly its dependence on assumptions such as model linearity and parameter accuracy. This realization led to increasing interest in exploring experimental design problems within the Bayesian framework. The Bayesian approach integrates prior information and model uncertainty with utility functions that describe experimental objectives, allowing for the selection of designs that maximize the expected utility of the experimental outcomes, following the theory of optimal decision-making under uncertainty.

Pioneering contributions to Bayesian optimal experimental design include seminal papers by Lindley in 1968 \citep{lindley1968choice} and 1972 \citep{lindley1972bayesian}, which highlighted the role of experimental objectives in design, and Chaloner's 1982 work \citep{chaloner1982optimal}, which extended the theory within the context of linear regression. Pilz's 1991 book \citep{pilz1991bayesian} further discussed the application of Bayesian methods for parameter estimation in linear regression models and considered the impact of parameter uncertainty on experimental design.
In recent years, simulation-based DOE methods have gained traction, particularly those utilizing Markov Chain Monte Carlo (MCMC) techniques \citep{clyde1995exploring, bielza1999decision, stroud2001optimal, muller2005simulation, amzal2006bayesian, muller2006bayesian, cook2008optimal}, adaptive design optimization based on mutual information \citep{cavagnaro2010adaptive}, and experimental design methods integrated with active learning\citep{ryan2015fully, toman1994efficiency, kadane2011bayesian, etzioni1993optimal, han2004bayesian}. These advances reflect a growing trend towards more flexible and computationally robust approaches to experimental design under uncertainty.

In 2023, Wu et al\citep{wu2023large}. introduced a novel approach leveraging derivative-informed projected neural networks for large-scale Bayesian optimal experimental design, particularly suited for optimizing experiments involving partial differential equations (PDEs).  However, a notable limitation of this approach is its dependency on the availability of accurate gradients, which can be problematic in scenarios with noisy or highly complex data. 
Aushev et al.\citep{aushev2023online} proposed an online simulator-based Bayesian design method tailored to complex models, particularly in fields like cognitive science.  This method offers greater flexibility and can be applied to a broad spectrum of models, its primary drawback lies in its high computational cost, especially when dealing with large-scale simulations.
Furthermore, in 2024, Valentin et al.\citep{valentin2024designing} applied machine learning techniques to optimize behavioral experiments using BOED. They presented a flexible workflow that integrates recent advancements in machine learning and simulation-based methods to optimize designs for complex psychological models. However, their approach faces limitations related to the necessary simplifications made to the models in order to ensure computational feasibility, which may lead to experimental designs that do not fully capture the theoretical intricacies of the behavior under study.
In 2024, Orozco R et al.\citep{orozco2024probabilistic} trained a scalable Conditional Normalizing Flow (CNF) to efficiently maximize the expected information gain for joint learning experimental design. However, their method incurs a relatively high computational cost, with numerical examples requiring up to 20 hours of training time on a single GPU.
There also exists a large body of related work\citep{foster2019variational,dong2025variational} that leverages variational methods or approximates the expected utility using lower or upper bounds. Due to space limitations, we do not elaborate on these approaches here.
These methods collectively illustrate the rich landscape of modern BOED research, showcasing diverse approaches such as gradient-informed neural networks, likelihood-free inference, and flow-based generative models. While each technique offers valuable contributions tailored to specific applications, they all face trade-offs between computational efficiency, model fidelity, and scalability. 
The primary challenges stem from the reliance on gradient information and the substantial computational demands associated with these methods.

To overcome these challenges, we utilize conditional density estimation to approximate the ratio of two Gaussian random fields under distinct constraints within a Bayesian framework. This approach is aimed at accelerating model fitting and integral computation, ultimately yielding the optimal design solution that maximizes the Bayesian posterior expectation integral. The key contributions of this work are as follows: 
\begin{itemize}
\item[•] A reformulation based on the ratio of two Gaussian random fields using independent Monte Carlo integration is proposed to reduce the computational complexity induced by the original nested integration.
\item[•] A machine learning-based conditional density estimation approach is employed to approximate the constraint density, thereby avoiding repeated surrogate modeling.
\item[•] A covariance-guided criterion is proposed to efficiently select informative datasets for both model training and Monte Carlo integration.
\item[•] The problems of failure probability estimation, parameter inference, and surrogate modeling are reformulated within the unified framework of Bayesian Optimal Experimental Design.
\end{itemize}

The structure of this paper is organized as follows. Section~\ref{sec: A basic framework for Bayesian optimization of experimental design} introduces the foundational concepts of Bayesian Optimal Experimental Design and reformulates the expected utility function using the Kullback–Leibler divergence and Bayes’ theorem. Section~\ref{sec:Gaussian process regression} presents Gaussian process regression, the key statistical modeling tool employed throughout the study. Section~\ref{sec:Constraint informed statistical modelling via conditional density learning}, which forms the core of this work, focuses on enhancing the computational efficiency of BOED. It leverages conditional density estimation to learn Gaussian random fields under varying constraints and utilizes the posterior covariance of the Gaussian process to identify informative datasets for statistical modeling and Monte Carlo integration. Section~\ref{set:PoI} describes how failure probability estimation, parameter inference, and surrogate modeling problems can be reformulated within the unified BOED framework. In Section~\ref{set:Numerical experiments}, a series of numerical examples—including failure probability estimation, parameter estimation, and efficient surrogate construction—are presented to demonstrate the feasibility and effectiveness of the proposed methodology. This section also includes a comparative analysis of the proposed design scheme against baseline methods such as random sampling and Latin Hypercube Design (LHD). Finally, Section~\ref{set:conclusion} concludes the paper with a summary of the main contributions and outlines potential directions for future research.

\section{Bayesian optimal experimental design}
\label{sec: A basic framework for Bayesian optimization of experimental design}

We begin by expressing the posterior expected integral of Bayesian experimental design using the principles of Bayesian inference. This formulation leads to the expression in Eq.~(\ref{eq12}), which serves as the core framework for the optimal design method proposed in this work.

\subsection{Bayesian optimal experimental design}

The goal of Bayesian optimal experimental design is to find a optimal design point $\d^*$ in the design space $D$ by maximizing the expectation utility function $U(\d,\bm{z} ,y)$ with respect to the model output (possible observation of simulation/experiment) $y$  of input $\d$ and parameter of interest (PoI) $\bm{z}$ \citep{huan2013simulation}, i.e.:
\begin{equation}\label{eq1}
\begin{aligned}
\d^{*} &= \underset{\d \in D}{\arg \max } U(\d, \bm{z}, y) \\
      &= \underset{\d \in D}{\arg \max } \int_{Y} \int_{\mathcal{X}} u(\d, \bm{z}, y) p(\bm{z}, y | \d) d \bm{z} d y \\
      &= \underset{\d \in D}{\arg \max } \int_{Y} \left(\int_{\mathcal{X}} u(\d, \bm{z}, y) p(\bm{z} | \d, y) d \bm{z}\right) p(y | \d) d y,
\end{aligned}
\end{equation}
where $\mathcal{X}$ and $Y$ denote the spaces of  $\bm{z}$ and $y$, respectively. 
The utility function \( u(\d, \bm{z}, y) \) serves to evaluate and compare potential experiments, directing the selection process towards choosing the experiment that maximizes expected utility. 
Additionally, \( p(\bm{z} | \d, y) \) represents the posterior distribution of the parameter of interest (PoI) \(\bm{z}\) given the experimental data, where the input is \(\d\) and the output is \(y\). 
Meanwhile, the distribution \( p(y | \d) \) represents the probabilistic output prediction for the design \(\d\).

Eq. \eqref{eq1} demonstrates that the optimal design within the Bayesian framework aims to maximize the posterior expected utility given the observations. Specifically, this corresponds to the term in parentheses in the third line of the middle equation of Eq.(\ref{eq1}). However, unless the likelihood and prior are carefully chosen to allow for analytical evaluation of the integral, Eq.(\ref{eq1}) typically lacks a closed-form solution. Consequently, numerical approximation techniques or stochastic solution methods are required to address both the optimization problem and the integral computation.

\subsection{The utility function}

In the realm of Bayesian optimal experimental design, the utility function plays a vital role. Selecting an appropriate utility function can facilitate the identification of more efficient design points and mitigate computational complexity. It is crucial, however, that this choice be informed by the specific objectives of the experiment and tailored to the parameter of interest. Here, we employ the Kullback-Leibler divergence (KLD), a widely utilized metric for measuring the difference between probability distributions, as our utility function. It's defined as follows:

\begin{equation*}\label{eq2}
    D_{KL}(p \| q) = \int p(x) \log \frac{p(x)}{q(x)} \, dx.
\end{equation*}
Within the framework of Bayesian experimental design, the KL divergence between the posterior and prior distributions serves as the utility function, denoted by \( u(\d, y) \). It is expressed as:

\begin{equation}\label{eq3} 
    u(\d, y) := D_{KL}\big(p(\bm{z} | \d, y) \| p(\bm{z})\big) = \int_{\bm{z}} p(\bm{z} | \d, y) \log \frac{p(\bm{z} | \d, y)}{p(\bm{z})} \, d\bm{z},
\end{equation}
where \( u(\d, y) \) quantifies the information gain about the parameter of interest (PoI) \(\bm{z}\) prior to and subsequent to conducting the experiment at the design point \(\d\) and observing its outcome \(y\).

\subsection{Reformulating the expected utility function} 

According to Eq.~\eqref{eq1}, the Bayesian experimental design framework endeavors to maximize the posterior expected utility function based on the observed data. 
This process constitutes an optimization problem, with an initial focus on addressing the double integral present in the objective term. The Kullback-Leibler (KL) divergence, as specified in Eq.~\eqref{eq3}, is employed as the utility function. Applying Bayes' theorem, the expression \(p\left(\bm{z} | \d,y\right)\) is reformulated as:
\begin{equation}\label{eq8}
    p\left(\bm{z} | \d, y\right) = \frac{p\left(y | \d, \bm{z}\right) p\left(\bm{z}\right)}{p\left(y | \d\right)}.
\end{equation}
Incorporating Eq.~\eqref{eq8} into Eq.~\eqref{eq3}, the utility function is subsequently expressed as:
\begin{equation}\label{eq9}
    u\left(y, \bm{z}, \d\right) = \int \frac{p\left(y | \d, \bm{z}\right) p\left(\bm{z}\right)}{p\left(y | \d\right)} \ln\frac{p\left(y | \d, \bm{z}\right)}{p\left(y | \d\right)} \, d\bm{z}.
\end{equation}
By substituting Eq.~\eqref{eq9} into Eq.~\eqref{eq1}, we derive a reformulated expression that relies solely on the prior distribution of the PoI \(\bm{z}\) and the observed data \(y\),
\begin{equation}\label{eq11}
\d^{*} = \arg\max_{\d \in D} \int_{Y} \int_{\bm{z}} \int_{\bm{z}} p\left(y | \d, \bm{z}\right) p\left(\bm{z}\right) \ln\frac{p\left(y | \d, \bm{z}\right)}{p\left(y | \d\right)} d\bm{z} \, p\left(\bm{z} | y, \d\right) d\bm{z} \, dy.
\end{equation}
Note that there are two integrals with respect to $\bm{z}$ in Eq.~(\ref{eq11}). The second integral with respect to $\bm{z}$ corresponds to a single full integral, as the integral of the posterior distribution $p(\bm{z} | y, \d)$ over $\bm{z}$ equals one. Therefore, Eq.~(\ref{eq11}) can be simplified and rewritten as:
\begin{equation}\label{eq12}
\d^{*} = \arg\max_{\d \in D} \int_{\bm{z}} \int_{Y} p(y | \d, \bm{z}) \ln\left(\frac{p\left(y | \d, \bm{z}\right)}{p\left(y | \d\right)}\right) dy \, p\left(\bm{z}\right) d\bm{z}.
\end{equation}

Eq.~(\ref{eq12}) depends solely on the prior distribution \( p(\bm{z}) \) of the PoI \( \bm{z} \), as well as the two conditional probabilities \( p(y | \d, \bm{z}) \) and \( p(y | \d) \), which are determined by the observed data \( y \). These quantities can be derived from the known conditions under which the surrogate model is constructed. Consequently, Eq.~(\ref{eq12}) takes the desired form of an optimal experimental design problem, where the relevant terms require optimization and approximation. The details of this process will be introduced in Section~\ref{sec:Constraint informed statistical modelling via conditional density learning}. Before that, however, we first describe how to model \( p(y | \d) \) and \( p(y | \d, \bm{z}) \)  using Gaussian process regression  in Section~\ref{sec:Gaussian process regression}.

\section{Gaussian process regression}
\label{sec:Gaussian process regression}

In this section, we elucidate the core principles of Gaussian Process Regression (GPR) and its application in modeling the distributions \( p(y | \d) \) and \( p(y | \d, \bm{z}) \). Given that \(\bm{z}\) is defined within the same domain as \(\d\), the conditional distribution \( p(y | \d, \bm{z}) \) is determined solely by the observation at the input \(\bm{z}\), denoted as \( y_{\bm{z}} \). These constraints, when embedded within the GPR framework, are equivalent  to incorporating additional data pairs \(\{ \bm{z}, y_{\bm{z}} \}\). For handling more intricate constraints, Gaussian Process Regression with Constraints (GPRC) \citep{wang2021explicit} offers a robust approach to effectively manage these complexities.

\subsection{Gaussian Process Regression}
Gaussian process regression (GPR) is a prominent regression model extensively utilized in statistical learning to address continuous variable regression problems. A distinctive feature of Gaussian processes is that their probability density functions are uniquely determined by their mean and covariance functions.

Formally, a Gaussian process is defined as a collection of random variables, any finite subset of which adheres to a multivariate Gaussian distribution. Specifically, a Gaussian process \( f(\d) \) is characterized by its mean function \( m(\d) \) and covariance function (or kernel) \( k(\d, \d') \):

\[
f(\d) \sim \mathcal{GP}(m(\d), k(\d, \d')).
\]

For simplicity, the mean function is typically assumed to be zero, i.e., \( m(\d) = 0 \). The covariance function encapsulates the relationships between points within the input space.

The selection of the covariance function is critical, as it dictates the smoothness, periodicity, and other properties of the functions modeled by the Gaussian process. Commonly employed kernels include:

\begin{itemize}
    \item \textbf{Radial Basis Function (RBF) Kernel}:
    \[
    k(\d_i, \d_j) = \sigma^2 \exp\left(-\frac{|\d_i - \d_j|_{L_2}}{2l^2}\right).
    \]
    This kernel assumes smoothness and is infinitely differentiable, making it suitable for modeling smooth functions.

    \item \textbf{Mat\'ern Kernel}:
    \[
    k(\d_i, \d_j) = \frac{2^{1 - \nu}}{\Gamma(\nu)} \left( \frac{\sqrt{2\nu} |\d_i - \d_j|}{l} \right)^\nu K_\nu\left( \frac{\sqrt{2\nu} |\d_i - \d_j|}{l} \right).
    \]
    The Mat\'ern kernel offers flexibility in controlling the smoothness of the resulting functions through the parameter \( \nu \), allowing for the modeling of functions with varying degrees of differentiability.
\end{itemize}

\subsubsection{Training and Prediction}

Consider a data-set \(\{(\d_i, y_i)\}_{i=1}^n\), where \( y_i \) represents the observed outputs. We assume a generative model for the data as follows:
\[
y_i = g(\d_i) + \epsilon_i.
\]
Here, \( g(\d) \) is an unknown latent function to be inferred, and \( \epsilon_i \sim \mathcal{N}(0, \sigma_n^2) \) denotes additive Gaussian noise with variance \( \sigma_n^2 \). This noise assumption implies that each observed output \( y_i \) deviates from the true function \( g(\d_i) \) by an amount characterized by Gaussian noise, thereby modeling the inherent uncertainty in the observations.

Assuming the latent function \( g(\d) \) follows a Gaussian process prior, any finite collection of function values across the input locations \( \mathbf{D} = \{\d_1, \dots, \d_n\} \) jointly adhere to a multivariate Gaussian distribution. Consequently, the observed outputs \( \mathbf{y} = [y_1, \dots, y_n]^\top \) also follow a multivariate Gaussian distribution:
\[
\mathbf{y} \sim \mathcal{N}(\mathbf{m}, \mathbf{K} + \sigma_n^2 \mathbf{I}),
\]
where \( \mathbf{m} \) is the mean vector of the Gaussian process evaluated at the training points \( \mathbf{D} \), and \( \mathbf{K} \) is the covariance matrix computed using the kernel function \( k(\d, \d') \), which encodes prior beliefs about the smoothness and other properties of \( f(\d) \). The term \( \sigma_n^2 \mathbf{I} \) accounts for observation noise by adding \( \sigma_n^2 \) to the diagonal of the covariance matrix, reflecting the assumption of independent noise across observations.

The marginal likelihood \( p(\mathbf{y} | \mathbf{D}) \) quantifies the probability of observing the data \( \mathbf{y} \) given the input locations \( \mathbf{D} \). Under the Gaussian assumptions, this marginal likelihood is expressed as a multivariate Gaussian distribution:
\[
p(\mathbf{y} | \mathbf{D}) = \mathcal{N}(\mathbf{y} | \mathbf{m}, \mathbf{K} + \sigma_n^2 \mathbf{I}).
\]
Maximizing this marginal likelihood with respect to the kernel hyperparameters facilitates the fitting of the Gaussian process model to the observed data in a principled manner. This process involves maximizing the log marginal likelihood:
\[
\log p(\mathbf{y} | \mathbf{D}) = -\frac{1}{2} \mathbf{y}^\top (\mathbf{K} + \sigma_n^2 \mathbf{I})^{-1} \mathbf{y} - \frac{1}{2} \log |\mathbf{K} + \sigma_n^2 \mathbf{I}| - \frac{n}{2} \log(2\pi).
\]
The log marginal likelihood comprises three components: the first term penalizes deviations of the observed outputs from the model's predicted mean, the second term penalizes model complexity through the log determinant of the covariance matrix, and the third term serves as a normalization constant. This balance between data fit and model complexity promotes a parsimonious model that mitigates the risk of overfitting.

For new input points \( \mathbf{D}^* = \{\d^*_1, \dots, \d^*_m\} \), we aim to determine the predictive distribution of the corresponding outputs \( \mathbf{y}^* = [y^*_1, \dots, y^*_m]^\top \). Within the Gaussian process framework, the joint distribution of the training outputs \( \mathbf{y} \) and the test outputs \( \mathbf{y}^* \) is given by:

\[
\begin{pmatrix}
\mathbf{y} \\
\mathbf{y}^*
\end{pmatrix} \sim \mathcal{N}\left(\begin{pmatrix}
\mathbf{m} \\
\mathbf{m}^*
\end{pmatrix}, \begin{pmatrix}
\mathbf{K} + \sigma_n^2 \mathbf{I} & \mathbf{K}_{*} \\
\mathbf{K}_{*}^\top & \mathbf{K}_{**}
\end{pmatrix}\right).
\]

In this formulation, \( \mathbf{m}_* \) denotes the mean vector for the test points, \( \mathbf{K}_{*} \) represents the covariance between the training and test points, and \( \mathbf{K}_{**} \) is the covariance matrix of the test points themselves. By conditioning on the observed data \( \mathbf{y} \), the posterior predictive distribution of \( \mathbf{y}^* \) is derived as:
\[
\mathbf{y}^* | \mathbf{D}^*, \mathbf{D}, \mathbf{y} \sim \mathcal{N}(\bm{\mu}_*, \bm{\Sigma}_*),
\]
where the predictive mean \( \bm{\mu}_* \) and covariance \( \bm{\Sigma}_* \) are defined by:
\[
\bm{\mu}_* = \mathbf{K}_{*}^\top (\mathbf{K} + \sigma_n^2 \mathbf{I})^{-1} \mathbf{y},
\]
\[
\bm{\Sigma}_* = \mathbf{K}_{**} - \mathbf{K}_{*}^\top (\mathbf{K} + \sigma_n^2 \mathbf{I})^{-1} \mathbf{K}_{*}.
\]
The predictive mean \( \bm{\mu}_* \) serves as the best estimate of the function values at the new input locations, while the predictive covariance \( \bm{\Sigma}_* \) quantifies the uncertainty associated with these predictions.

Gaussian Process Regression offers a robust probabilistic framework for regression tasks, adept at capturing uncertainty in predictions and facilitating the flexible modeling of complex functions. The efficacy of GPR hinges on the judicious selection of the covariance function, which embodies assumptions regarding the smoothness and variability of the underlying function. 

\section{Enhancing Modeling Efficiency}
\label{sec:Constraint informed statistical modelling via conditional density learning}

The nested integral in Equation \eqref{eq22} exacerbates the computational burden of utility expectation.  To address this challenge, we propose a novel transformation of \( p(y|\d,\bm{z}) \) as $p(y|\d) \frac{p(y|\d,\bm{z})}{p(y|\d)}.$ With this approach, the BOED framework can be reformulated as an independent integral.
\begin{equation}
\label{eq:10}
\d^{*} = \arg\max_{\d \in D} \int_{\bm{z}} \int_{Y} \frac{p\left(y | \d, \bm{z}\right)}{p\left(y | \d\right)} \ln\left(\frac{p\left(y | \d, \bm{z}\right)}{p\left(y | \d\right)}\right) p(y | \d)p\left(\bm{z}\right) dy d\bm{z}.
\end{equation}

In comparison to Equation \eqref{eq12}, Equation \eqref{eq:10} still involves a double integral, but the variables \( \z \) and \( y \) are independent. Double integrals with independent variables are more computationally efficient, particularly when large-scale sampling is involved, as they significantly reduce the computational burden.

We address the significant computational complexity arising from repeatedly modeling Gaussian process regression surrogates when the conditional density \( p(y|\d,\bm{z}) \) is unknown. To alleviate this computational burden, we propose an approximation for the density \( p(y|\d,\bm{z}) \), denoted as \( \hat{p}(y|\d,\bm{z}) \), and defined as:

\begin{equation}
\label{eq:cde_app}
\hat{p}(y|\d,\bm{z}) = \mathcal{Z}q_{CDE}(y|\d,\bm{z})p(y|\d) \propto q_{CDE}(y|\d,\bm{z})p(y|\d),
\end{equation}
where \( \mathcal{Z} \) is a normalizing constant, and \( q_{CDE}(y|\d,\bm{z}) \) serves as an approximation for the ratio \( \frac{p(y|\d,\bm{z})}{p(y|\d)} \) using the conditional density estimation method.

There are two compelling reasons for learning \( q_{CDE}(y|\d,\bm{z}) \) instead of directly modeling the constraint Gaussian process \( p(y|\d,\bm{z}) \). Firstly, the distribution of \( \frac{p(y|\d,\bm{z})}{p(y|\d)} \) is inherently simpler than that of \( p(y|\d,\bm{z}) \). This simplicity arises from the fact that the output prediction at \( \d \), under the influence of constraints, decreases with an increasing distance between \( \z \) and \( \d \).
Secondly, the training process can be expedited by selecting an informative training set, a process elaborated in detail in Section \ref{set:inform_data}.

\subsection{Approximation by Conditional Density Estimation} 

In Eq.~(\ref{eq:10}), the terms \( p(y|\d) \) and \( p(y|\d,\bm{z}) \) represent the density functions of observed data \( y \), conditioned on \( \d \) and subject to various constraints \( \bm{z} \). To explicitly define these density functions for calculation, we apply two types of Gaussian Process Regression:

\begin{itemize}
    \item Standard GPR is utilized to model \( p(y|\d) \), yielding \( p_{\text{GP}}(y|\d) \).
    \item Constrained GPR  is employed to model \( p(y|\d,\bm{z}) \), resulting in \( p_{\text{GP}}(y|\d,\bm{z}) \). The constraint is modeled as a virtual observation with an additional pair of data, denoted as \(\{\bm{z}, y_{\bm{z}}\}\).
\end{itemize}  

These approaches correspond to two Gaussian random fields with almost same level of complexity. 
With the use of Gaussian process surrogates, we know that both \( p(y|\d) \) and \( p(y|\d, \bm{z}) \) follow Gaussian distributions. Considering the properties of Gaussian distributions, the ratio of two Gaussian densities is also an unnormalized Gaussian density. Specifically, we have:
\[
\frac{\mathcal{N}(y | \mathbf{m}_1, \Sigma_1)}{\mathcal{N}(y | \mathbf{m}_2, \Sigma_2)} = \mathcal{N}(y | \mathbf{m}, \Sigma) \cdot \mathcal{Z},
\]
where
\begin{equation}
\label{eq:ratio_GP}
\Sigma = \left(\Sigma_1^{-1} - \Sigma_2^{-1}\right)^{-1}, \quad \mathbf{m} = \Sigma \left( \Sigma_1^{-1} \mathbf{m}_1 - \Sigma_2^{-1} \mathbf{m}_2 \right),
\end{equation}
and
\[
\mathcal{Z} = \frac{\left|\Sigma_2\right|}{\left|\Sigma_2 - \Sigma_1\right|} \cdot \frac{1}{\mathcal{N}(\mathbf{m}_1 | \mathbf{m}_2, \Sigma_2 - \Sigma_1)}.
\]
Thus, the ratio \( \frac{p(y|\d, \bm{z})}{p(y|\d)} \) in  Equation~(\ref{eq12}) can be approximated by the product of  conditional density \( q_{CDE}(y|\d, \bm{z}) \) and function $\mathcal{Z}(\d,z)$,
We estimate this ratio using the conditional density estimation technique, specifically employing the Kernel Mixture Network (KMN) introduced in Section \ref{set:cde}, yielding:
Substituting Eq.~(\ref{eq:cde_app}) into Eq.~(\ref{eq:10}), we have
\begin{equation}\label{eq14}
   \d^{*} = \arg\max_{\d \in D} \int_{\bm{z}} \int_{Y} \frac{ \hat{p}(y|\d,\bm{z})}{p\left(y | \d\right)} \ln\left(\frac{ \hat{p}(y|\d,\bm{z}) }{p\left(y | \d\right)}\right) p(y | \d)p\left(\bm{z}\right) dy d\bm{z}.
\end{equation}
And a computational expression with nested Monte Carlo form is 
\begin{equation}\label{eq15}
   \d^* = \arg\max_{\d \in D} \mathbb{E}_{y_i\sim p_{GP}(y|\d),\z_i\sim p(z)} \frac{ \hat{p}(y_i|\d,\bm{z}_i)}{p\left(y_i | \d\right)}  \ln(\frac{ \hat{p}(y_i|\d,\bm{z}_i)}{p\left(y | \d\right)} ).
\end{equation}

Furthermore, once the training process of the density \( q_{CDE}(y|\d,\z) \) is completed, subsequent calls to \( q_{CDE}(y|\d,\z) \) incur negligible computational cost. As a result, the Monte Carlo integration in Equation \eqref{eq15} is expected to be highly efficient.

\subsection{Kernel Mixture Network for Conditional Density Estimation}
\label{set:cde}

Let \((\bm{d},\bm{z}, y)\) be a pair of random variables in the spaces \(D\),   \(\mathcal{X}\) and \(\mathcal{Y}\), respectively. 
The conditional probability distribution \(p(y|\bm{d},\bm{z}) = \frac{p(y,\bm{d},\bm{z})}{p(\bm{d},\bm{z})}\) is defined, where \(p(y,\bm{d},\bm{z})\) represents the joint density function of \(y\) and \(\bm{d},\bm{z}\), and \(p(\bm{d},\bm{z})\) is the marginal density function of \(\bm{d},\bm{z}\). This formulation facilitates the derivation of the conditional distribution of \(y\) given specific values of \(\bm{d},\bm{z}\).

Kernel Mixture Networks (KMN) \citep{ambrogioni2017kernel} provide a robust framework for conditional density estimation by effectively combining the flexibility of kernel methods with the expressive power of neural networks. The KMN approach excels at modeling complex conditional distributions by using mixtures of kernel functions to approximate the underlying density.

\paragraph{Architecture Overview}

The architecture of a typical KMN consists of several key components:
\begin{itemize}
    \item \textbf{Input Layer}: This layer accepts the explanatory variable \(\bm{d},\bm{z}\), which can be a vector of features or a scalar input.
    \item \textbf{Hidden Layers}: A series of hidden layers (usually fully connected) process the input, capturing nonlinear relationships in the data. Common activation functions such as ReLU or sigmoid introduce non-linearity to the model.
    \item \textbf{Output Layer}: The output layer generates the parameters of the kernel mixture, including the weights, means, and bandwidths that define the mixture density.
    \item \textbf{Kernel Functions}: KMNs typically utilize Gaussian kernels, which are defined as follows,
    \[
    K(\bm{d},\bm{z}, \mu_i, h_i) = \frac{1}{\sqrt{2\pi} h_i} \exp\left(-\frac{(\bm{d},\bm{z} - \mu_i)^2}{2 h_i^2}\right).
    \]
    Here, \(\mu_i\) denotes the mean and \(h_i\) represents the bandwidth of the \(i\)-th kernel.
\end{itemize}

The conditional density \(p(y|\bm{d},\bm{z})\) is estimated as a weighted sum of the individual kernel densities:

\[
\hat{p}(y|\bm{d},\bm{z}) = \sum_{i=1}^K w_i K(y, \mu_i, h_i)
\]

In this expression, \(K\) denotes the number of kernels in the mixture, \(w_i\) is the weight of the \(i\)-th kernel (ensuring that \(\sum_{i=1}^K w_i = 1\)), and \(K(y, \mu_i, h_i)\) represents the kernel function centered at \(\mu_i\) with bandwidth \(h_i\).

Training the KMN involves optimizing the network parameters through a suitable loss function. Typically, the negative log-likelihood of the observed data under the estimated conditional density is minimized. Optimization techniques such as stochastic gradient descent (SGD) or Adam are commonly employed for this task.
The loss function is defined as:
\[
\mathcal{L} = -\sum_{j=1}^N \log\left(\hat{p}(y_j|\bm{d}_j,\bm{z}_j)\right),
\]
where \((\bm{d}_j,\bm{z}_j, y_j)\) represent the observed pairs of data points.

\subsection{Covariance-Driven Acceleration Method}

In the numerical computation of Equation \eqref{eq15}, the CDE modeling and the integral evaluation of the utility function are computationally intensive tasks. To address this, we propose a covariance-based acceleration method that selects the  informative data-set for CDE training and Monte Carlo integration, thereby reducing the computational cost associated with both CDE learning and Monte Carlo simulations.
This approach leverages the constraint definition: \( \z \), which lies in the same space as the design \( \d \), and \( y_z \), which resides in the same space as \( y_d \). Consequently, we can effectively assess the contribution of the prediction \( p(y|\d,\z) \), which is influenced by the constraints \( \z \).
Both \( p(y|\d) \) and \( p(y|\d,\z) \) are modeled using Gaussian process regression, a kernel-based covariance modeling technique. By employing a Euclidean distance-based kernel function, the covariance decreases as the distance between inputs increases.
For a given design \( \d \), the distribution \( p(y|\d,\z) \) converges to \( p(y|\d) \)—which does not depend on the constraint at \( \z \)—when the covariance between \( y_d \) at design \( \d \) and \( y_{z_i} \) at location \( \z_i \) becomes negligible.
An intuitive interpretation of this is that both \( p(y|\d) \) and \( p(y|\d,\z) \) are modeled by Gaussian process regression, where the covariance is influenced by the Euclidean distance between the inputs. When the covariance between the design and the constraint approaches zero, $p(y|\d,\z)$ reduces to $p(y|\d)$.

\subsubsection{ Informative data-set construction for CDE learning}
\label{set:inform_data}
In the training process of the conditional distribution \( p_{CDE}(y|\d,\z) \), the data-set \( \mathcal{D}_{CDE} = \{\d_i,\z_i, y_i\}_{i=1}^n \) plays a crucial role. However, evaluating \( y \) at design \( \d \) with constraint $\z$ is computationally expensive due to the iterative modeling process required. To optimize both the accuracy and efficiency of the model, a highly informative data-set is essential. 
Incorporating the knowledge of a given non-uniform prior \( p(\z) \) during each iteration enables the model to effectively prioritize high-probability regions within the domain. By integrating this prior information, we guide the search process towards regions of the parameter space that are more likely to provide meaningful insights. The locations of the constraints are determined by the structure of the prior, and areas with higher density values are more likely to contribute valuable information to the model, thereby enhancing the overall efficiency and accuracy of the learning process. This approach not only ensures that the model focuses on the most promising regions, but also facilitates the identification of important features in the data, which might otherwise be overlooked when relying solely on uniform priors.
Consequently, a joint distribution for generating an informative data-set is proposed. This data-set is denoted as \( \mathcal{D}_{inf} = \{\d_i, \z_i \}_{i=1}^n \), where \( \d_i, \z_i \sim p(\d|\z) p(\z) \). 

The conditional distribution \( p(\d|\z) \) is defined as a uniform distribution over a limited support \( \Omega_z \), where:

\begin{equation}
\label{eq:infor_z}
\Omega_z := \{ \d \in \mathcal{X} |  \text{COV}(y_d, y_z) \geq \epsilon_{cov} \},
\end{equation}
where \( \text{COV}(\cdot, \cdot) \) represents the covariance function of the Gaussian process regression model \( p(y|\d) \), and \( \epsilon_{cov} \) is a threshold for selecting the design points for training. This threshold ensures that the selected data points are those where the correlation between the design variables \( \d \) and the constraints \( \z \) is sufficiently strong.

The rationale behind this approach is that training data with higher correlation between design variables and constraints leads to a model that better captures the influence of \( \z \) on \( y_d \). In contrast, sample pairs where \( \text{COV}(y_d, y_z) < \epsilon_{cov} \) provide limited information, as the prediction \( p(y|\d,\z) \) will approach \( p(y|\d) \), where the constraint effect \( \z \) is negligible or absent. In this case, the conditional density \( p_{CDE}(y|\d,\z) \) will converge to a constant, independent of both \( \d \) and \( \z \), resulting in a less informative contribution to the model. 

Thus, by focusing on regions of high correlation, we ensure that the model training is both efficient and informative, allowing for more precise and constraint-aware predictions.
The specific computational procedure is provided in the Algorithm \ref{alg:infor_data-set}.

\begin{algorithm}[H]
\caption{Informative data-set Construction}
\label{alg:infor_data-set}
\begin{algorithmic}[1]
\REQUIRE 
  GPR model $p(y|\d)$;
  Distribution of PoI $p(\z)$;
  
\ENSURE $\mathcal{D}_{\mathrm{cde}}$

\STATE Construct the sample set $Z = \{\z_i\}_{i=1}^{N_{\z}} \sim p(\z)$ by MCMC.
    \STATE Construct the subset $Z_1 \gets \{\z_i\}_{i=1}^{n_{\z}}$ by selecting $n_{\z}$ samples from $Z$.
    \STATE Initialize $\mathcal{D}_{\mathrm{cde}} \gets \emptyset$
    \FOR{each $\z_i \in Z_1$}
        \STATE Select $n_d$ candidate designs $\{\d_{ij}\}_{j=1}^{n_d} \subset \mathcal{D}$ using Eq.\eqref{eq:infor_z}.
        \FOR{ $\tilde{\d} = \d_{i1}\dots,\d_{in_d}$}
            \STATE Predict $(m_{1,ij}, \Sigma_{1,ij}), (m_{2,ij}, \Sigma_{2,ij}) \gets p_{\mathrm{GP}}(y|\tilde{\d}, \z_i), p_{\mathrm{GP}}(y|\tilde{\d})$.
            \STATE Obtain $(\hat{m}_{ij}, \hat{\Sigma}_{ij})$ using Eq.\eqref{eq:ratio_GP}.
            \STATE Sample $\hat{y}_{ij} \sim \mathcal{N}(\hat{m}_{ij}, \hat{\Sigma}_{ij})$.
            \STATE Update $\mathcal{D}_{\mathrm{cde}} \gets \mathcal{D}_{\mathrm{cde}} \cup \{\tilde{\d}, \z_i, \hat{y}_{ij}\}$.
        \ENDFOR
    \ENDFOR
\end{algorithmic}
\end{algorithm}

\subsubsection{Informative samples selection of Monte Carlo integral}
\label{set:effic_doubleMC}

The problem of Bayesian optimal experimental design is inherently an optimization problem, where the primary objective is to maximize  the utility function, as expressed in Equation \eqref{eq15}. 
The central task in this context is to identify the optimal design point \( \d_* \). 
Once the model \( q_{CDE}(y|\d, \bm{z}) \) has been trained, the computational burden primarily arises from the calculation of expectations. 

In the Monte Carlo integration, for each fixed design point \( \d_i \), it is necessary to compute a summation of the form \( q_{CDE}(y|\d_i,\z_j) \ln q_{CDE}(y|\d,\z_j) \) over all samples drawn from the prior distribution, i.e., $\z_j\sim p(\z)$. 
Notably, at this stage, the design point \( \d_i \) is already fixed, and thus, not all samples drawn from \( p(\z) \) contribute to the prediction of \( y \). Specifically, if the correlation between the predicted responses \( y_{\z_j} \) and the fixed design point \( y_{\d_i} \) is small, the prediction for \( y_{\d_i} \) will not be significantly affected by the corresponding constraints. 
In such cases, the value of \( q_{CDE}(y|\d_i,\z_j) \) remains constant, and no substantial information gain is obtained from those samples.

To mitigate this inefficiency, we can exploit the covariance \( \text{COV}(y_{\d_i}, y_{\z_j}) \) between the predicted response at the fixed design point \( \d_i \) and the response at the samples \( \bm{z}_j \) in $p(y|\d)$ model. 
By introducing the tolerance threshold \( \epsilon_{cov} \), we can filter out those \( \bm{z} \) values that exhibit high covariance with the fixed design point.,
\begin{equation}\label{eq16}
    \{\bm{z} | \operatorname{COV}(y_{\d_i}, y_{\bm{z}})>\epsilon_{cov}, \bm{z}\sim p(\bm{z}) \}.
\end{equation}
This filtering step effectively reduces the number of samples that need to be considered, improving the computational efficiency of the Monte Carlo estimation while retaining the relevant information for the optimization process. 
In summary, this strategy enhances the overall efficiency of the Bayesian experimental design process by eliminating irrelevant or redundant samples, thus ensuring that the optimization process is both computationally feasible and accurate.
The specific computational procedure is provided in the Algorithm. \ref{alg:samples_MC}.

\begin{algorithm}[H]
\caption{ Informative Samples Selection for Monte Carlo integral}
\label{alg:samples_MC}
\begin{algorithmic}[1]
\REQUIRE 
  design candidate $\d_i$;
  
\ENSURE $Z_{d_i}$
  
\STATE Initial informative sample set $Z_{d_i}  \gets \emptyset$
        \FOR{each $\z \in Z$}
            \STATE Compute the covariance  prediction between $\d_i$ and $\z$ by GPR.
            \IF {$\text{COV}(y_{\d_i},y_{\z})>\epsilon_{cov}$}
            	\STATE $Z_{d_i} = Z_{d_i}\cup \{ \z\}$
            \ENDIF
        \ENDFOR
\end{algorithmic}
\end{algorithm}

The adoption of covariance as a screening parameter \( \epsilon_{cov} \) offers two key benefits: first, it accelerates model fitting during CDE approximation; and second, it reduces computational complexity in numerical integration.
\section{Experimental design scheme} 
\label{sset:Experimental design scheme}

\subsection{Definition of PoI in surrogate modelling, parameter estimation and failure probability estimation}
\label{set:PoI}
\textbf{Surrogate Modelling}\\
The primary focus of studying the efficient surrogate model construction problem within the framework of Bayesian design of experiments is to address the regions of large variance in the model. Let the true function of the model be denoted as \( g(\bm{x}) \). The goal of the efficient surrogate model construction problem is to approximate \( g(\x) \) with a model \( f(\x) \), with particular emphasis on optimizing the high variance regions of \( f(\x) \).
Therefore, the definition of PoI in surrogate modelling can be defined by the function of variance of current probabilistic surrogate, as following

\begin{equation}\label{eq22}
  p\left(\z\right){:}=p\left(z|\psi\circ\sigma\left(\z\right)\right),
\end{equation}
where $\psi(\cdot)$ denotes a positive function and $\sigma(\z)$ is the variance of probabilistic surrogate.
In the Gaussian Process Regression (GPR) model fitted to the data-set \( (\mathbf{X}, \bm{y}) \), we can sample the high variance regions of the model as \( p(\z) \) and optimize them by searching for the optimal experimental design points. This optimization aims to enhance the predictive performance of the surrogate model. The stopping criterion for this optimization process is that the optimal experimental design points can no longer lead to a significant reduction in the root-mean-square error (RMSE).\\

\textbf{Parameter estimation}\\
The key point in studying the parameter estimation problem in the framework of Bayesian design of experiments lies in obtaining the posterior distributions of the model parameters or sampling their distributions, defining the true posterior distribution of the parameter estimation problem as:
\begin{equation}\label{eq20}
    p\left(\z\right){:}=p\left(\bm{z}|y=y_0\right).
\end{equation}
Then, in the GPR model fitted with data-sets $(\mathbf{X},\bm{y})$, we can sample an approximation of the true posterior distribution $\hat{p}\left(\bm{z}|y=y_0\right)$, and similarly, as the model is fitted more accurately, the approximation of the posterior distribution $\hat{p}\left(\bm{z}|y=y_0\right)$ will gradually converge to the true posterior distribution $p\left(\bm{z}|y=y_0\right)$. The design of experiments reduces the uncertainty in the model predictions and improves the accuracy of the fit. The stopping condition of the experimental design is that the experimental design point no longer affects the posterior distribution of the model samples.\\

\textbf{Failure Probability  estimation}\\
The key point in studying the problem of estimating the probability of failure in the framework of design of experiments is to obtain the experimental points near the failure boundary or the approximate failure boundary.We define the state function of the failure probability problem as $g(\x)$, and its failure boundary is defined as $\ell{:}\left\{\x\in\Omega\left|g\left(\x\right)=0\right\}\right.$, then define
\begin{equation}\label{eq17}
    p\left(\z\right):=p\left(\z|g\left(\cdot\right)\right)=\frac1C\mathrm{exp}\left(-\frac{\left|g\left(\z\right)-0\right|}\lambda\right)
\end{equation}

This is called the limiting state distribution of the failure boundary, where C is the regularization constant, $\lambda$ is the scale parameter, $\z$ is the point near the failure boundary, and $\lambda$ is used to control the bandwidth, , to control the distance of point $\z$  from the true failure boundary. It is well understood that, since Eq.(\ref{eq17}) are obviously approximating the distribution $g(\z)=0$ of sampled , the corresponding point $\z$ should fall near $g(\z)=0$, i.e., the distribution of $\z$ will be clustered at the failure boundary, but since the true state function is not known, i.e., the failure boundary is not known, and in the case of inefficient simulation and costly data acquisition,  Eq.(\ref{eq17}) are obtained by sampling the surrogate model GPR. 

In failure probability estimation problem,  the stopping condition can be defined as $\left|\frac{\hat{P}_{t+1}-\hat{P}_t}{\hat{P}_t}\right|<\varepsilon $ , where $\hat{P}_t$ is the probability of failure for the $t$-th iteration.

\subsection{Numerical algorithm}

The methodological workflow of the accelerated Bayesian optimal experimental design proposed in this study is formally outlined in Algorithm \ref{alg:design_optimization}. Notably, the stopping criterion defined in Line 10 is inherently adaptable, contingent on the specific objectives of different experimental design problems. 
For instance, in conventional experimental design applications, the primary objective often revolves around maximizing predictive accuracy, wherein the process concludes once the deviation between model predictions and true values falls within an acceptable threshold. Conversely, in applications such as failure probability estimation, the algorithm terminates when the selected sampling points sufficiently approximate the true failure boundary. This flexibility extends to parameter estimation and various other research domains, ensuring that the framework accommodates a broad spectrum of experimental objectives while maintaining computational efficiency.

\begin{algorithm}[H]
\caption{Accelerated Bayesian Optimal Experiment Design (Acc-BOED)}
\label{alg:design_optimization}
\begin{algorithmic}[1]
\REQUIRE 
  Size of initial data-set $n_{\mathrm{in}}$; 
  Convergence threshold $\epsilon_z$;
  
\ENSURE GPR model $p(y|\d)$
  
\STATE Construct the initial data-set $\mathcal{D}=\{\mathbf{D},  \mathbf{y}\}$ by $n_0$ design points $\mathbf{D}=[\d_1,\dots,\d_{n_0}]^\top$ decided by LHS and $\mathbf{y} = [y(\d_1),\dots,y(\d_{n_0})]^\top$ which is the corresponding outputs by taking simulations.
\STATE Train GPR model and obtain $p_{\mathrm{GP}}(y|\d)$ on the data-set $\mathcal{D}$.
\STATE Define distribution of the PoI random variable $\z$ reffed in Section \ref{set:PoI} and draw the samples from the defined $p_0(\z)$ by MCMC method with the given GPR model.
\FOR{$t=1,\dots,n_{max}$}
    \STATE Construct the data-set $\mathcal{D}_{CDE}$ as Algorithm.\ref{alg:infor_data-set} and Train the CDE model $q_{\mathrm{CDE}}(y|\d,\z)$ on the data-set.
    \STATE Find the optimal design $\d_t^*$ by maximizing Eq. \eqref{eq15} with the relation $\hat{p}(y|\d,\z) = q_{CDE}(y|\d,\z)p_{\mathrm{GP}}(y|\d)$ and the Informative samples in Algorithm.\ref{alg:samples_MC}.
    \STATE Obtain experimental measurement $y(\d_t^*)$ and update data-set $\mathcal{D}$.
    \STATE Retrain the GP model $p_{\mathrm{GP}}(y|\d)$ 
    \IF{ Meet specific stopping criteria }
    	\STATE Break.
    \ENDIF
    \STATE Update the distribution of PoI $p_t(\z)$ and draw the samples.
\ENDFOR
\end{algorithmic}
\end{algorithm}

\section{Numerical experiments}
\label{set:Numerical experiments}

In this section, we apply the Accelerated Bayesian Optimal Experimental Design (Acc-BOED) Algorithm~\ref{alg:design_optimization} to tackle three specific challenges: efficient surrogate model construction, parameter estimation, and failure probability estimation. Our primary objective is to assess the feasibility and effectiveness of the Acc-BOED framework. To benchmark its performance, we perform a comparative analysis against two widely utilized sampling techniques: Latin Hypercube Sampling (LHS) and random point sampling.

The effectiveness of the proposed experimental design framework in addressing the previously mentioned challenges is clearly demonstrated in this study. 
A significant finding is the substantial computational acceleration achieved by the Acc-BOED method in comparison to the standard Bayesian Optimal Experimental Design (BOED) approach. Table~\ref{tab:time_comparison} provides a summary of the execution time, measured in seconds, necessary for completing a full iteration—from dataset selection to identifying the Bayesian optimal experimental design—across six test cases pertaining to the three identified problems. The first row of the table details the computation time required by Acc-BOED, the second row outlines the time for the basic BOED method, and the third row illustrates the relative time efficiency. As indicated by the results in Table~\ref{tab:time_comparison}, Acc-BOED consistently surpasses the basic method, achieving a speedup factor between 6 to 13 times over the six test cases. This highlights that Acc-BOED can markedly reduce computational costs while maintaining high precision in experimental results.

\begin{table}[ht]
\centering
\small
\caption{Execution Time Comparison (seconds)}
\label{tab:time_comparison}
\begin{tabularx}{\textwidth}{@{}c *{6}{>{\centering\arraybackslash}X}@{}} 
\toprule
\multirow{4}{*}{Method} & \multicolumn{6}{c}{Test Cases} \\
\cmidrule(lr){2-7}
& \multicolumn{2}{c}{\textit{Surrogate Model}} 
& \multicolumn{2}{c}{\textit{Parameter Estimation}} 
& \multicolumn{2}{c}{\textit{Failure Probability}} \\
\cmidrule(lr){2-3} \cmidrule(lr){4-5} \cmidrule(lr){6-7}
& Trig. Model & Alanine Model & Gauss Function & KDV Equ. & Circle Problem & Four Branch \\
\midrule
Acc-BOED & 321 & 346 & 205 & 138 & 274 & 193 \\
Basic BOED & 2765 & 2308 & 2231 & 1874 & 3251 & 1234 \\
\textbf{Ratio (Basic/Acc)} & \textbf{8.61} & \textbf{6.67} & \textbf{10.88} & \textbf{13.58} & \textbf{11.87} & \textbf{6.39} \\
\bottomrule
\end{tabularx}

\smallskip
\begin{minipage}{\textwidth}
\footnotesize
\textit{Note:} Abbreviations: Trigonometric Model (Trig. Model), Alanine Dipeptide Model (Alanine Model), KDV Equation (KDV Equ.), Four Branch System (Four Branch).
\end{minipage}
\end{table}

\subsection{Efficient Surrogate Model}

\subsubsection{Trigonometric Model}
We begin with a trigonometric function defined on a two-dimensional domain \( \Omega = (-4, 8) \times (-4, 8) \). The true function \( f: \Omega \to \mathbb{R} \), which serves as the target model for our surrogate modeling problem, is defined as
\begin{equation}
\label{eq:true_function}
    f(\mathbf{x}) = \sin(x_1) \cdot \cos(x_2),
\end{equation}

where \( \mathbf{x} = (x_1, x_2) \in \Omega \) denotes the input vector. The primary objective is to construct an accurate and computationally efficient surrogate model over this domain.

The efficient surrogate modeling problem begins with an initial dataset of 30 points, sampled using Latin hypercube sampling (LHS). To improve the predictive accuracy of the surrogate model, Algorithm~\ref{alg:design_optimization} is applied to compute the root mean square error (RMSE) of the current surrogate model, where \( p(\theta) \) is determined by Eq.~\eqref{eq22}. The Bayesian optimal experimental design is then identified, and new design points are added to the dataset. The RMSE of the updated model is recomputed, and this process is iterated until RMSE convergence is achieved. After 20 iterations, the final dataset is expanded to 49 points. To evaluate the effectiveness of the Acc-BOED method, we conduct a comparative analysis against two alternative sampling strategies: random sampling and Latin hypercube sampling (LHS). In both cases, the dataset is similarly expanded from 30 to 49 points, and the RMSE of the Gaussian process regression (GPR) model with identical hyperparameters is compared.

\begin{figure}[H]
    \centering
    \begin{minipage}[b]{0.45\linewidth}
        \centering
        \includegraphics[width=\linewidth]{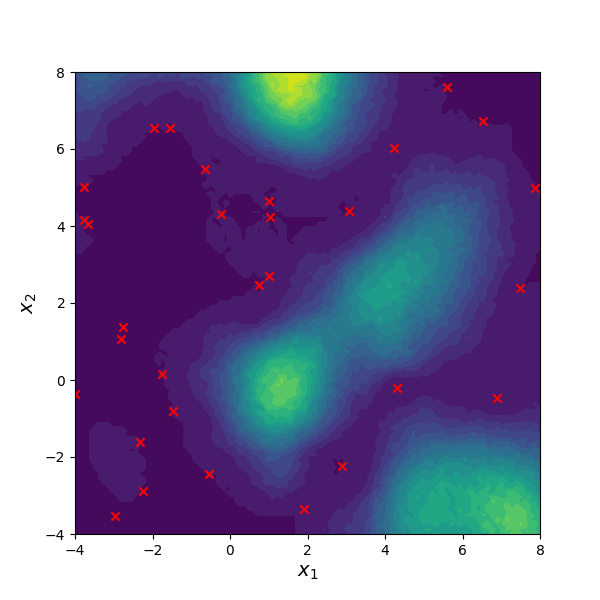}
        \subcaption{\scriptsize Expected Information Gain (EIG) of the Acc-BOED Method}
    \end{minipage}%
    \hfill
    \begin{minipage}[b]{0.45\linewidth}
        \centering
        \includegraphics[width=\linewidth]{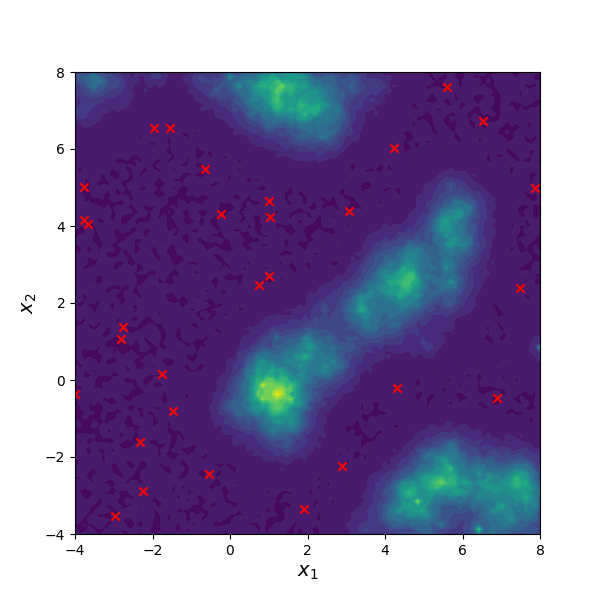}
        \subcaption{\scriptsize  Expected Information Gain (EIG) of the BOED Method}
    \end{minipage}
    
    \caption{\scriptsize(a) The left plot illustrates the values of \( u(d) \) for experimental design points obtained using Acc-BOED.  (b) The right plot presents the values of \( u(d) \) for experimental design points obtained using BOED}
    \label{fig:comparison2}
\end{figure}

Before presenting the comparative experimental results, the effectiveness of the Acc-BOED method is assessed, particularly its capability to accurately evaluate the expected information gain (EIG) of experimental designs, as defined by \( u(d) \) in Eq.~\eqref{eq3}. The values of \( u(d) \) computed using the Acc-BOED method are compared with those obtained using the basic BOED approach, as shown in Figure~\ref{fig:comparison2}. Here, Figure~\ref{fig:comparison2} illustrates the comparison of \( u(d) \) values for a dataset size of 30. The results reveal a high degree of similarity between the two plots, indicating that the numerical values remain consistent. This consistency confirms the effectiveness and feasibility of the Acc-BOED method in identifying optimal experimental designs.

\begin{figure}[H]
    \centering
    \begin{minipage}[b]{0.45\linewidth}
        \centering
        \includegraphics[width=\linewidth]{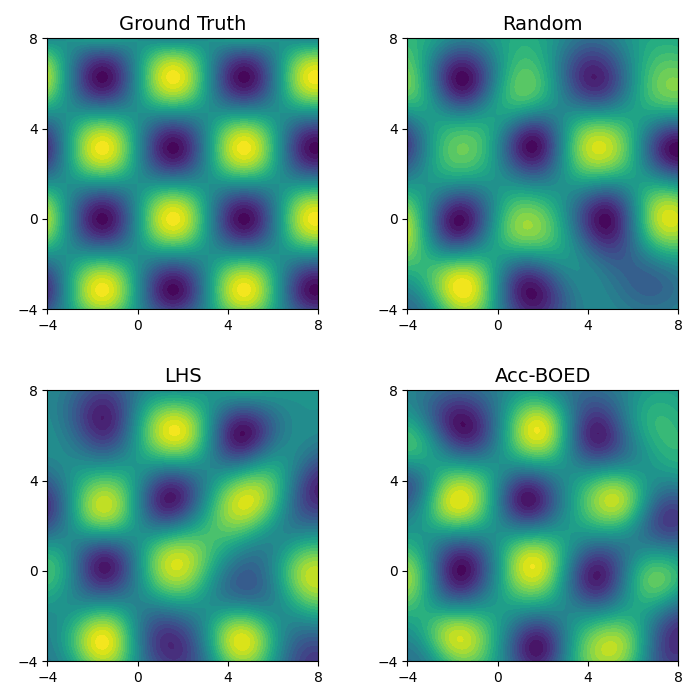}
        \subcaption{\scriptsize Comparison of final model predictions}
    \end{minipage}%
    \hfill
    \begin{minipage}[b]{0.45\linewidth}
        \centering
        \includegraphics[width=\linewidth]{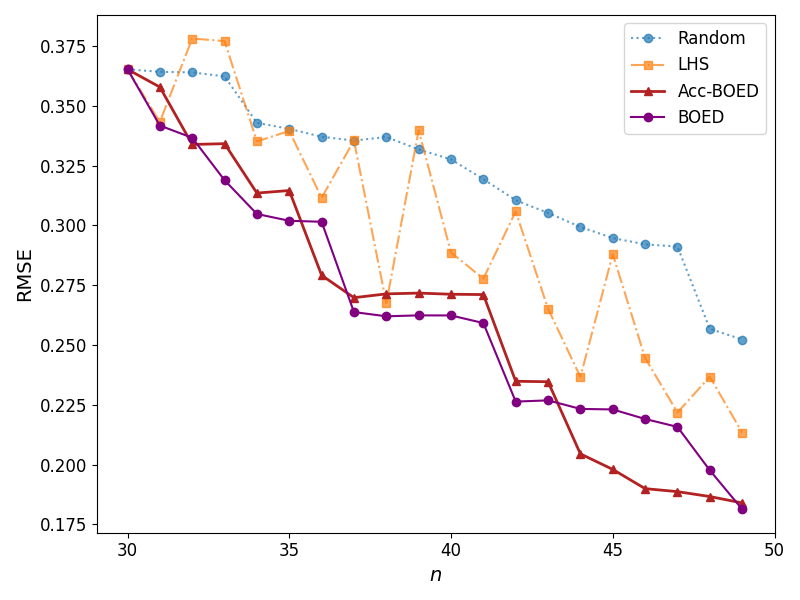}
        \subcaption{\scriptsize Evolution of RMSE}
    \end{minipage}
    
    \caption{\scriptsize (a) The left plot shows a comparison of the ground truth (top left) with model predictions obtained from 49 data points: LHS design (bottom left), random sampling (top right), and Acc-BOED (bottom right). (b) The right plot illustrates the evolution in the root mean square error (RMSE) as the number of data points increases from 30 to 49, comparing the performance of random sampling (blue), LHS  (yellow), Acc-BOED (red), and BOED (purple)}
    \label{fig:comparison}
\end{figure}

The model predictions of the three methods against the ground truth function at 49 data points are compared, as shown in Figure~\ref{fig:comparison}(a). It can be observed that the predictions of the Acc-BOED method align closely with the ground truth, demonstrating superior performance compared to both LHS and random sampling methods. The evolution of RMSE over 20 iterations is illustrated in Figure~\ref{fig:comparison}(b), where the horizontal axis represents the dataset size and the vertical axis denotes the RMSE value. The results show that the RMSE curve for the Acc-BOED method decreases significantly faster and ultimately outperforms both LHS and random sampling methods. Moreover, the RMSE curve of Acc-BOED remains nearly identical to that of the BOED method, indicating that Acc-BOED achieves computational acceleration without altering the optimal experimental design. This further validates the effectiveness of the Acc-BOED approach.

\begin{figure}[H]
    \centering
    \begin{minipage}[b]{0.9\linewidth}
        \centering
        \includegraphics[width=\linewidth]{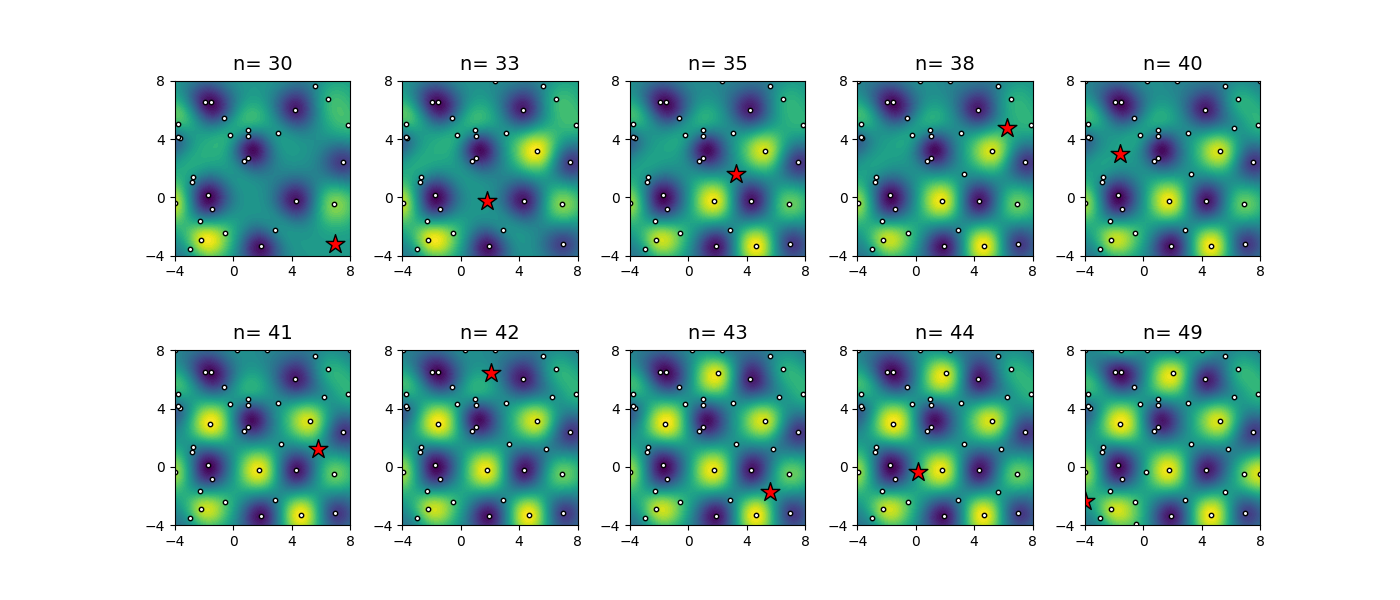}
        \subcaption{\scriptsize  Evolution of model prediction}
    \end{minipage}
    
    \vspace{0.5cm} 
    
    \begin{minipage}[b]{0.9\linewidth}
        \centering
        \includegraphics[width=\linewidth]{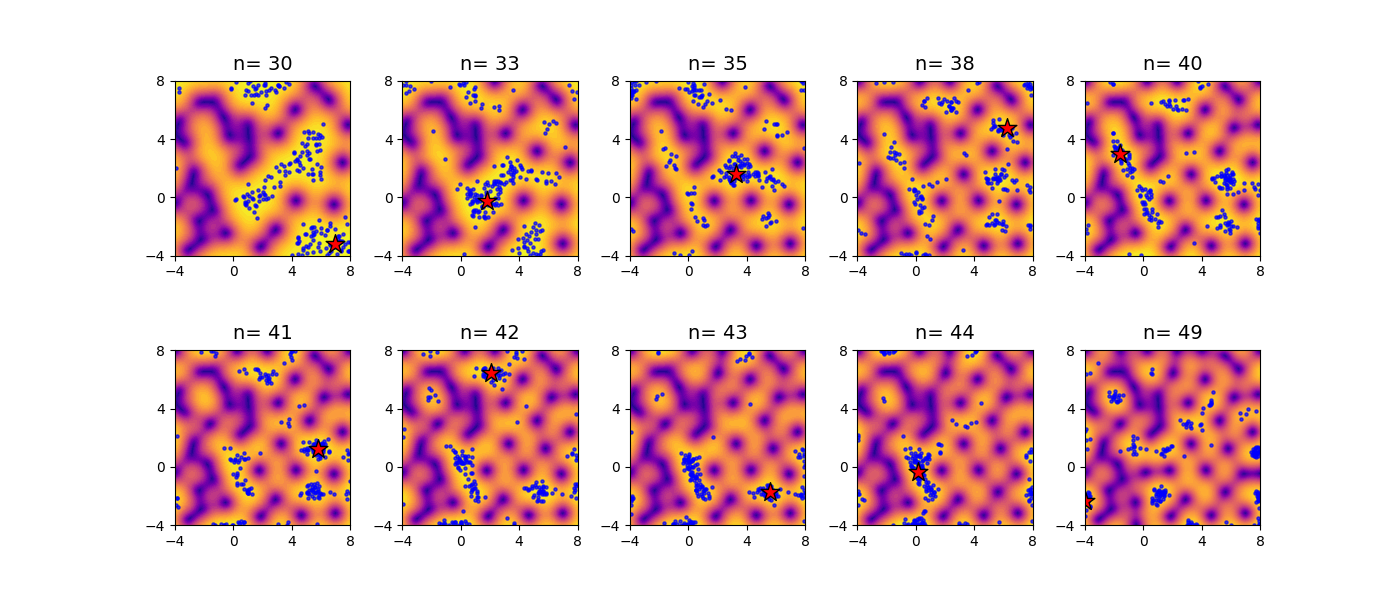}
        \subcaption{\scriptsize  Evolution of model variance}
    \end{minipage}
    
    \caption{\scriptsize (a) The first plot on the top illustrates the model predictions across 10 iterations, with the scatter points representing the data points at each iteration. (b) The plot on the bottom depicts the evolution of the model variance, where the scatter points correspond to the samples of \( p(\theta) \) as defined in Eq.~\eqref{eq22}, and the red star denotes the optimal experimental design point $d^*$ at each iteration}
    \label{fig:mean and var}
\end{figure}

The evolution of the Gaussian process regression (GPR) model predictions and their mean values over 20 iterations is illustrated in Figure~\ref{fig:mean and var}, with 10 selected iterations shown. In Figure~\ref{fig:mean and var}(a), the white dots represent the current data points, while in Figure~\ref{fig:mean and var}(b), the blue dots denote partial samples of \( p(\theta) \) as defined by Eq.~\eqref{eq22}, and the red star indicate the optimal experimental design point identified by the current model. From Figure~\ref{fig:mean and var}, it is evident that the experimental design points tend to concentrate in regions of high model variance, significantly improving the predictive accuracy of the model. These results demonstrate the feasibility and effectiveness of applying the Bayesian optimal experimental design framework to the construction of efficient surrogate models.

\subsubsection{Alanine Dipeptide Model}
In this study, we demonstrate the application of our proposed method to alanine dipeptide, a molecular dynamics model characterized by 22 dimensions and two torsion angles as its collective variables. Specifically, we investigate the isomerization process of alanine dipeptide in a vacuum environment at a temperature of \( T = 300 \) K. Due to its well-documented behavior in prior theoretical and computational research \citep{li2019computing, ren2005transition}, alanine dipeptide serves as an ideal benchmark for validating our approach.

\begin{figure}[H]
  \centering
  \includegraphics[width=0.5\linewidth]{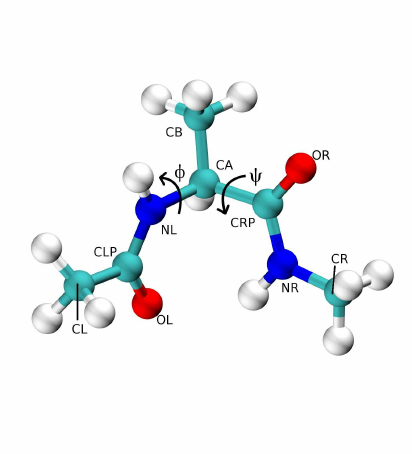}  
  \caption{Schematic representation
 of the alanine dipeptide(CH3
CONH–CHCH3–CONH–CH3)} 
  \label{fig:SM2model}
\end{figure}

The alanine dipeptide molecule, composed of 22 atoms, exhibits key features typical of biomolecules despite its relatively simple chemical structure. Figure~\ref{fig:SM2model} illustrates the molecule using a stick-and-ball representation, emphasizing the two torsion angles, \(\phi(\boldsymbol{x})\) and \(\psi(\boldsymbol{x})\), which serve as the collective variables. These variables are defined as functions of \(\boldsymbol{x}\), the Cartesian coordinates of all atoms in the system. The free energy associated with the collective variables \((\phi(\boldsymbol{x}), \psi(\boldsymbol{x}))\) is expressed as a function of \(\boldsymbol{\nu} = (\nu_1, \nu_2)\):

\begin{equation}
\label{sm2}
F(\boldsymbol{\nu}) = -k_B T \ln \left( A^{-1} \int e^{-\frac{V(\boldsymbol{x})}{k_B T}} \times \delta(\nu_1 - \phi(\boldsymbol{x})) \times \delta(\nu_2 - \psi(\boldsymbol{x})) \, d\boldsymbol{x} \right),
\end{equation}

where \( A = \int_{\mathbb{R}^d} e^{-\frac{V(\boldsymbol{x})}{k_B T}} \, d\boldsymbol{x} \) is a normalization factor, \( T \) denotes the temperature, \( k_B \) is the Boltzmann constant, \( \delta(\cdot) \) represents the Dirac delta function, and \( V(\boldsymbol{x}) \) is the potential energy function of the atomic positions \( \boldsymbol{x} \in \mathbb{R}^d \).

To construct the free energy landscape of the alanine dipeptide model, restrained simulations are performed to sample the Gibbs distribution \( e^{-V(\boldsymbol{x})/k_B T} \) while fixing the two torsion angles \(\boldsymbol{\nu} = (\phi, \psi)\), as described in Eq.~\eqref{sm2}. The free energy data \( F \) at specific angles are generated using the NAMD software package \citep{phillips2005scalable}, which simulates Langevin dynamics with a time step of 0.5 fs. The resulting numerical approximations of \( F \), which include inherent noise, are subsequently used in our method to annotate data points across the two-dimensional angle plane.

\begin{figure}[H]
  \centering
  \begin{minipage}[b]{0.45\textwidth}  
    \centering
    \raisebox{-\height}{  
      \includegraphics[width=\linewidth]{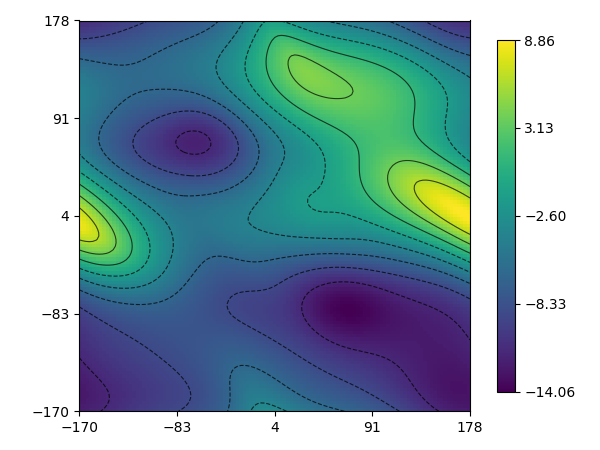}
    }
    \caption{The Ground Truth of Alanine Dipeptide Model} 
    \label{fig:SM2true}
  \end{minipage}%
  \hspace{0.5cm}  
  \begin{minipage}[b]{0.45\textwidth}  
    \centering
    \raisebox{-\height}{  
      \includegraphics[width=\linewidth]{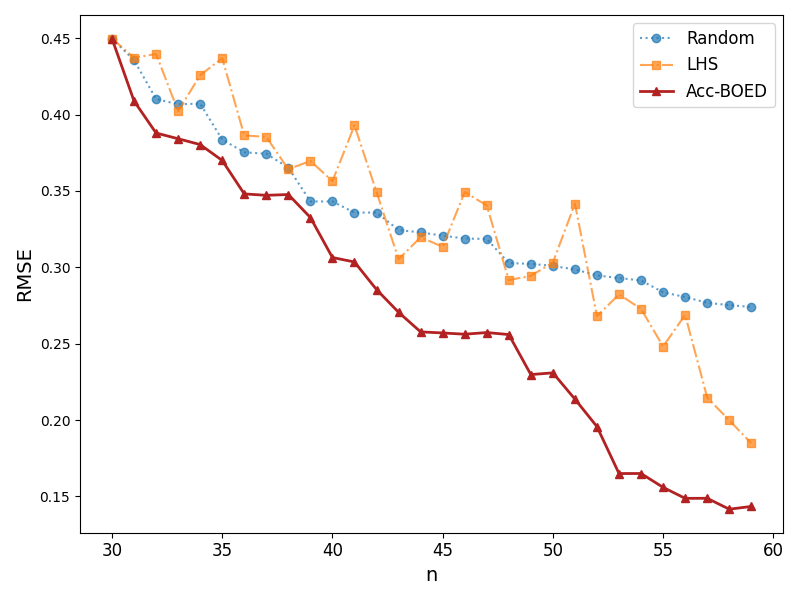}
    }
    \caption{Evolution of RMSE }
    \label{fig:sm2rmse}
  \end{minipage}
\end{figure}

For the Alanine Dipeptide model, we first compute the free energy data \( F \) at grid points corresponding to specific torsion angles, obtaining the true model as depicted in Figure~\ref{fig:SM2true}. The horizontal and vertical axes represent the values of the collective variables, while the heatmap indicates the corresponding Gibbs free energy \( F \). To construct an efficient surrogate model, we apply Algorithm~\ref{alg:design_optimization} to this problem, starting with an initial dataset of 30 points generated using Latin Hypercube Sampling (LHS). The root mean square error (RMSE) is iteratively updated until convergence is achieved. After 30 iterations, the final dataset is expanded to 59 points. Similarly, we compare the performance of random sampling and Latin Hypercube Sampling (LHS) by fixing the Gaussian Process Regression (GPR) hyperparameters and computing the RMSE of the GPR model for datasets ranging from 30 to 59 points.

\begin{figure}[H]
  \centering	\includegraphics[width=.8\linewidth]{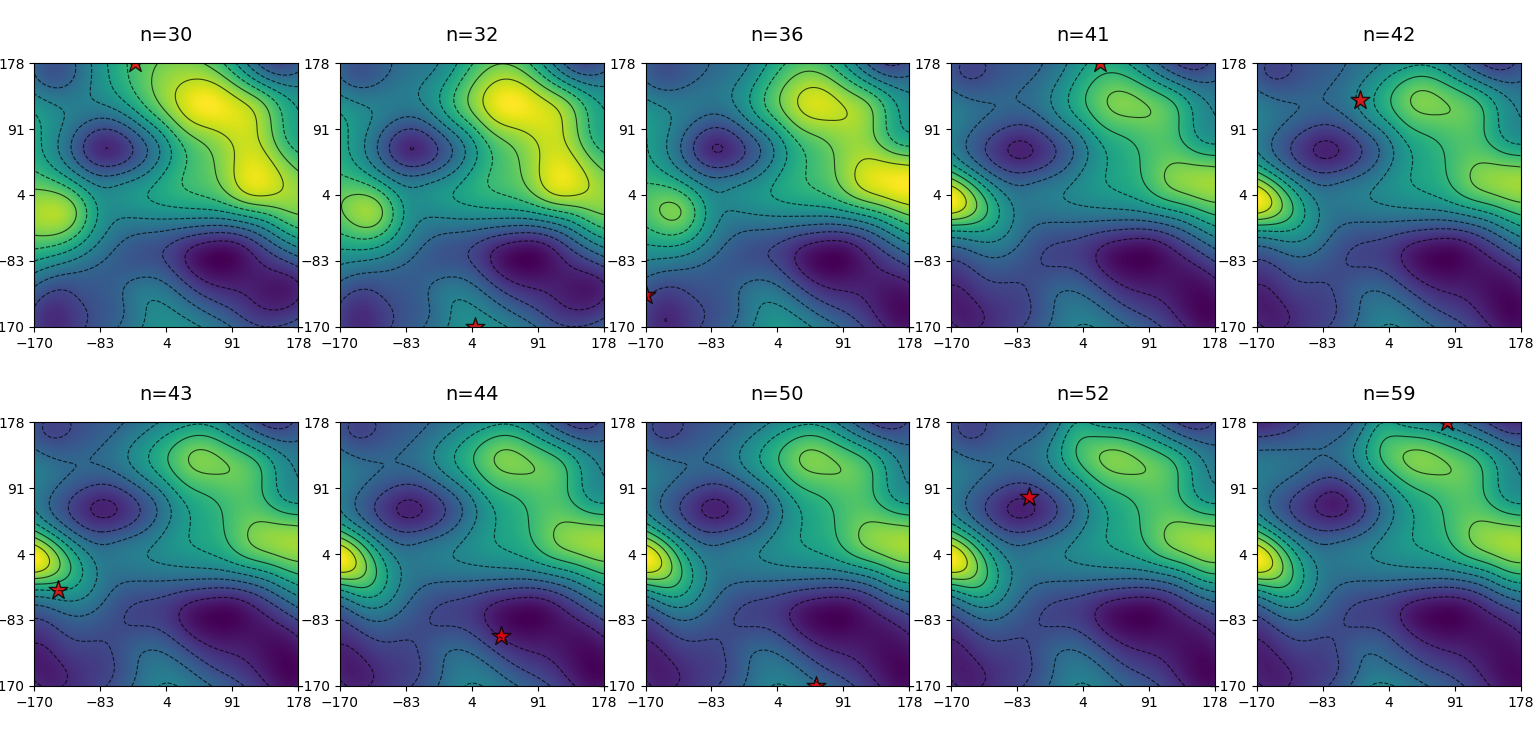}
	\caption{Evolution of ADM prediction} 
	\label{fig:mean10}
\end{figure}

The RMSE evolution over 30 iterations for random sampling, LHS approach, and the Acc-BOED method, with the number of data points increasing from 30 to 59, is shown in Figure~\ref{fig:sm2rmse}. It can be observed that the RMSE curve of the Acc-BOED method significantly outperforms the other two methods, stabilizing around 0.15 earlier, indicating that the model fitting accuracy is already very high. This demonstrates the superiority and reliability of the Acc-BOED approach. The corresponding model predictions for the selected ten iterations are illustrated in Figure~\ref{fig:mean10}, where it is evident that the predicted model gradually converges toward the true model over these iterations.

\subsection{Parameter estimation}
The key to studying parameter estimation within the framework of Bayesian Optimal Experimental Design lies in obtaining the posterior distribution of the model parameters. The true posterior distribution for the parameter estimation problem is defined as:
\begin{equation}\label{eq20}
    p\left(\theta\right) := p\left(\theta|y=y_0\right),
\end{equation}
where \(\theta\) represents the parameters of interest, and \(y_0\) denotes the observed data. After fitting a Gaussian Process Regression (GPR) model to the dataset \((D, y)\), we can sample the posterior distribution of the model using Bayes' theorem:
\begin{equation}
\label{bys}   
         p(\theta | y) \propto p(\theta) \cdot p(y | \theta),
\end{equation}
where \(p(\theta)\) is the prior distribution of \(\theta\), and \(p(y | \theta)\) is the likelihood function. As more information from the observed data \(y_0\) is incorporated, the approximate posterior distribution \(\hat{p}\left(\theta|y=y_0\right)\) becomes increasingly accurate and converges to the true posterior distribution \(p\left(\theta|y=y_0\right)\). The role of experimental design is to obtain a more accurate and reliable model posterior with as few observed data points \(y_0\) as possible. The experimental design process stops when the addition of new design points no longer significantly alters the posterior distribution of the model parameters.

In the one-dimensional scenario, the observed data corresponding to the true parameter \(\theta^*\) is a scalar value \(y^*\). A Gaussian Process Regression (GPR) model is fitted to the dataset \((\theta, y)\), and the approximate posterior distribution \(\hat{p}(\theta|y=y^*)\) is sampled using Bayes' theorem (Eq.~\eqref{bys}) within the fitted GPR framework. This approximate posterior is subsequently compared to the true posterior distribution \(p(\theta|y=y^*)\), which is also sampled using Bayes' theorem (Eq.~\eqref{bys}) but computed directly from the true model rather than the surrogate GPR model.

In the two-dimensional case, the observed data corresponding to the true parameter \(\theta^*\) is similarly a scalar \(y^*\), defined as the mean squared error (MSE) with respect to the condition with true parameter, ideally equal to zero. However, due to inherent noise in the data, \(y^*\) typically attains a small, non-zero value. The true posterior distribution \(p(\theta|y=y^*)\) is first sampled using Bayes' theorem (Eq.~\eqref{bys}) based on the true model. Subsequently, a surrogate GPR model is fitted to the dataset \((\theta, y)\), and the approximate posterior distribution \(\hat{p}(\theta|y=y^*)\) is sampled from the GPR.

In both scenarios, Algorithm~\ref{alg:design_optimization} is employed to iteratively identify the optimal experimental design until the posterior distributions converge. The performance metric utilized is the Kullback-Leibler (KL) divergence between the approximate posterior distribution and the true posterior distribution.

\subsubsection{Gaussian error function}

We introduce the Gaussian error function as the first example for the parameter estimation problem, defined as follows:
\begin{equation}\label{eq21}
\text{erf}(x)=\frac{1}{\sqrt{\pi}}\int_{-x}^x e^{-t^2} dt = \frac{2}{\sqrt{\pi}}\int_0^x e^{-t^2} dt
\end{equation}
The true model is defined by \( y = \text{erf}(\theta + \varepsilon) \), where \( \varepsilon \sim N(0, \sigma^2) \), \( \theta \in (0, 1) \), and \( \sigma = 0.1 \). The prior distribution of \(\theta\), \( p(\theta) \), is assumed to be uniform over the interval \((0, 1)\).

The parameter estimation problem requires estimating the true parameter \(\theta^*\) from the given observed data \( y^* = 0.6927 \), where the true value of \(\theta^*\) is 0.7.

\begin{figure}[H]
  \centering
	\includegraphics[width=.8\linewidth]{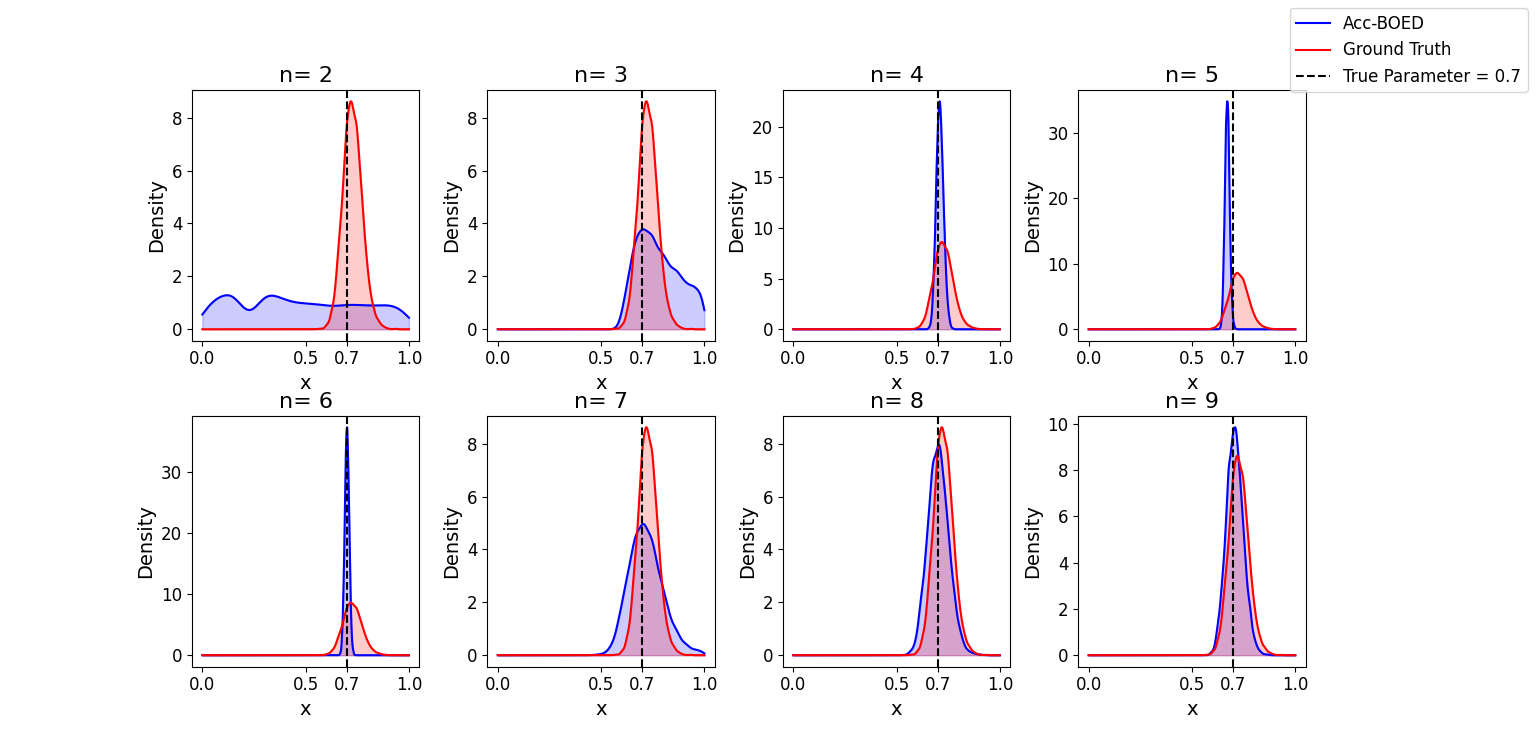}
	\caption{ Evolution of Approximate vs. True Posterior Distributions} 
	\label{fig:erf}
\end{figure}

Starting from an initial condition with a single data point, the optimal experimental design is iteratively refined over eight iterations. The final approximate posterior distribution, as shown in Figure~\ref{fig:erf}, converges to the true parameter value of 0.7. In the figure, the red distribution denotes the true posterior distribution, while the blue distribution represents the approximate posterior distribution. The black solid line marks the true parameter value \(\theta^*\). As clearly illustrated, the approximate posterior distribution progressively aligns with the true posterior distribution and becomes increasingly concentrated around \(\theta^*\). This simple one-dimensional example provides a compelling demonstration of the feasibility of the experimental design framework in solving parameter estimation problems.

\subsubsection{KDV equation}

Using a simple Gaussian error function for parameter estimation demonstrates the feasibility of Acc-BOED but does not fully highlight its superiority. In this section, we employ the Korteweg-de Vries (KdV) equation for parameter estimation and compare the performance of Acc-BOED with alternative methods, such as Latin Hypercube Sampling (LHS) and random sampling, to showcase its advantages.

The KdV equation is a nonlinear, dispersive partial differential equation for the function \( u \), which depends on space \( x \in \mathbb{R} \) and time \( t \):

\[
\frac{\partial u}{\partial t} + \theta_1 u \frac{\partial u}{\partial x} + \theta_2 \frac{\partial^3 u}{\partial x^3} = 0,
\]

This equation serves as an asymptotic simplification of the Euler equations and is commonly used to model shallow water waves. It can also be interpreted as the Burgers' equation augmented with an additional dispersive term. In this study, the parameters \( \theta_1 \) and \( \theta_2 \) are estimated from noisy observations of \( u \). The true solution is generated using a spectral method with 100 spatial points and 500 timesteps.

Due to the high nonlinearity of the PDE, 200 solution values are randomly selected from the time domain \([0, 5]\) and the spatial domain \([0, 30]\). The true parameter values are \( \theta_1^* = 6 \) and \( \theta_2^* = 1 \). The prior distributions for these parameters are defined as:

\[
\pi_\mathrm{prior}(\theta_1) \sim \mathcal{U}(3, 12), \quad \pi_\mathrm{prior}(\theta_2) \sim \mathcal{U}(0, 4).
\]

Gaussian noise with zero mean and standard deviation \( \sigma_\mathrm{obs} = 0.005 \) is added to the selected solution values.

\begin{figure}[H]
  \centering
  \begin{minipage}[b]{0.48\linewidth}
    \centering
    \includegraphics[width=\linewidth]{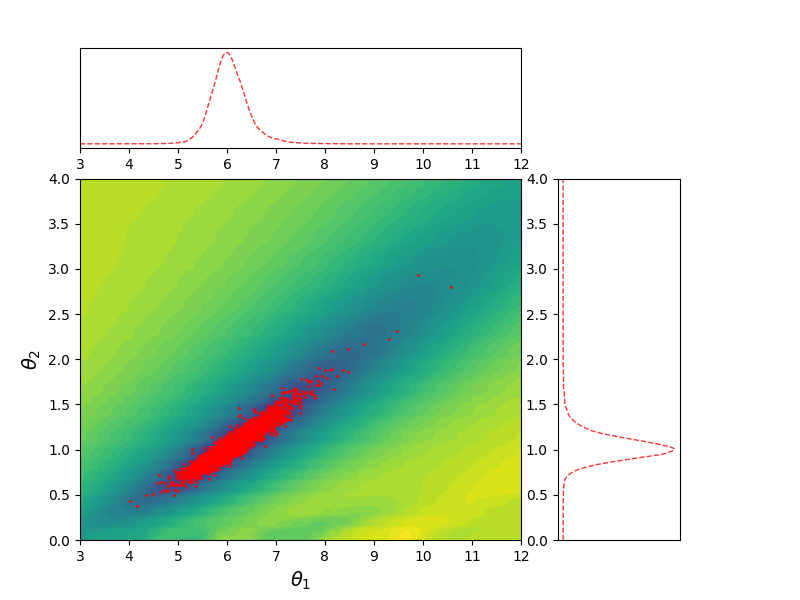}
    \caption{Parameter Sampling for Ground Truth in the KdV Equation}
    \label{fig:pereal}
  \end{minipage}%
  \hfill
  \begin{minipage}[b]{0.48\linewidth}
    \centering
    \includegraphics[width=\linewidth]{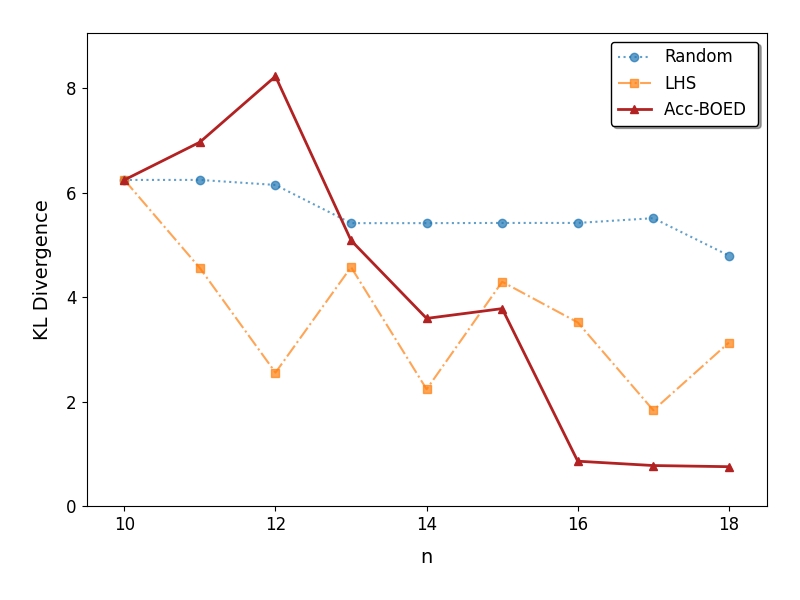}
    \caption{Evolution of KL Divergence Across Three Methodologies}
    \label{fig:KL10-18}
  \end{minipage}
\end{figure}

First, the true posterior distribution of the model is sampled according to Eq.~\eqref{bys}. The true model and its corresponding posterior distribution of parameters are shown in  Figure~\ref{fig:pereal}. It can be observed that the solution values are concentrated around the true parameters (6, 1), and the distributions of both parameters exhibit a Gaussian shape. Starting from an initial condition of ten data points, eight iterations of optimization are performed to search for the optimal experimental design. 

\begin{figure}[H]
    \centering
    \begin{minipage}[b]{0.23\linewidth}
        \centering
        \includegraphics[width=\linewidth]{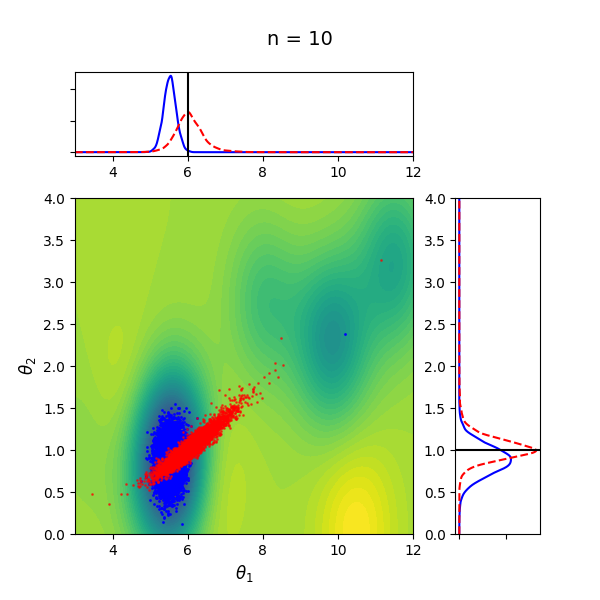}
    \end{minipage}%
    \hfill
    \begin{minipage}[b]{0.23\linewidth}
        \centering
        \includegraphics[width=\linewidth]{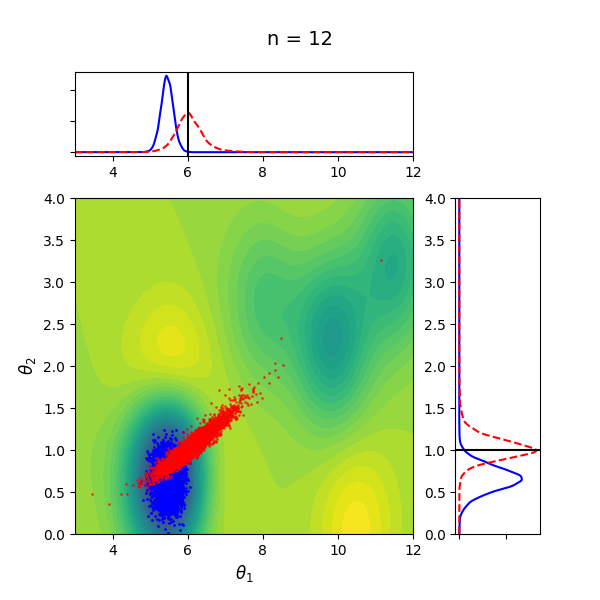}
    \end{minipage}%
    \hfill
    \begin{minipage}[b]{0.23\linewidth}
        \centering
        \includegraphics[width=\linewidth]{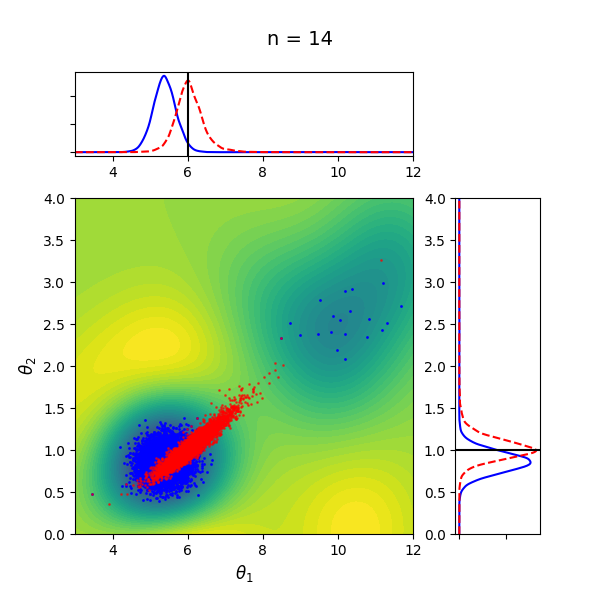}
    \end{minipage}%
    \hfill
    \begin{minipage}[b]{0.23\linewidth}
        \centering
        \includegraphics[width=\linewidth]{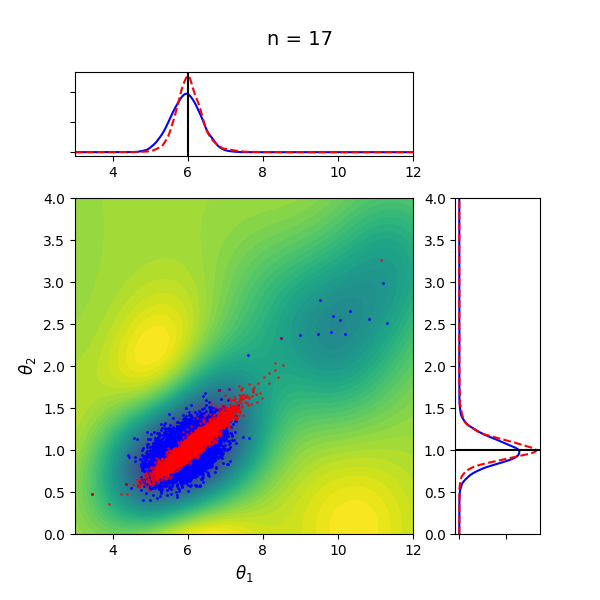}
    \end{minipage}

    \caption{\scriptsize Samples of approximate posterior distribution with respect to iterations: true posterior distribution (red scatter), approximate posterior distribution (blue scatter), and black solid line (true parameter point (6, 1))}
    \label{fig:pe10-18}
\end{figure}

The evolution of the model's Kullback-Leibler (KL) divergence is illustrated in Figure~\ref{fig:KL10-18}. Compared to LHS and Random sampling, the KL divergence of Acc-BOED decreases significantly faster and eventually converges to approximately 0.8, demonstrating the superiority of Acc-BOED. The sampling results of the approximate posterior distribution of the model at several iterations are shown in  Figure~\ref{fig:pe10-18}. It can be seen that the distributions of both parameters are ultimately concentrated near the true parameters. Although there is still a certain gap compared to the true posterior distribution, the performance of parameter estimation is expected to be very satisfactory.

\subsection{Failure Probability Estimation}

\subsubsection{Circle problem}

We begin by investigating the feasibility and effectiveness of transforming the failure probability estimation problem into a Bayesian optimal experimental design problem through a simple one-dimensional example.

The problem is formulated as follows: within a two-dimensional domain $(-10, 10)$, the true state function of the system is defined as
\begin{equation}\label{Circle}
    g(x_1, x_2) = 12 - x_1^2 - x_2^2,
\end{equation}
where $ x_1 $ and $ x_2 $ are the inputs to the system, independently and identically distributed according to $ N(0, 1) $, and $ y = g(x_1, x_2) $ represents the corresponding state function value.

In this problem, as indicated by Eq.~\eqref{Circle}, the true failure boundary is analytically represented as a circle. However, in practical engineering scenarios, the limit state function (i.e., Eq.~\eqref{Circle}) is typically a black-box function whose explicit form is unknown. Consequently, it becomes essential to model and simulate the true limit state function to approximate the failure boundary effectively. Given the inherent challenges of low computational efficiency in experimental simulations and the high costs associated with data acquisition, an experimental design approach is adopted to iteratively approximate the true failure boundary.

\begin{figure}[H]
  \centering
	\includegraphics[width=.8\linewidth]{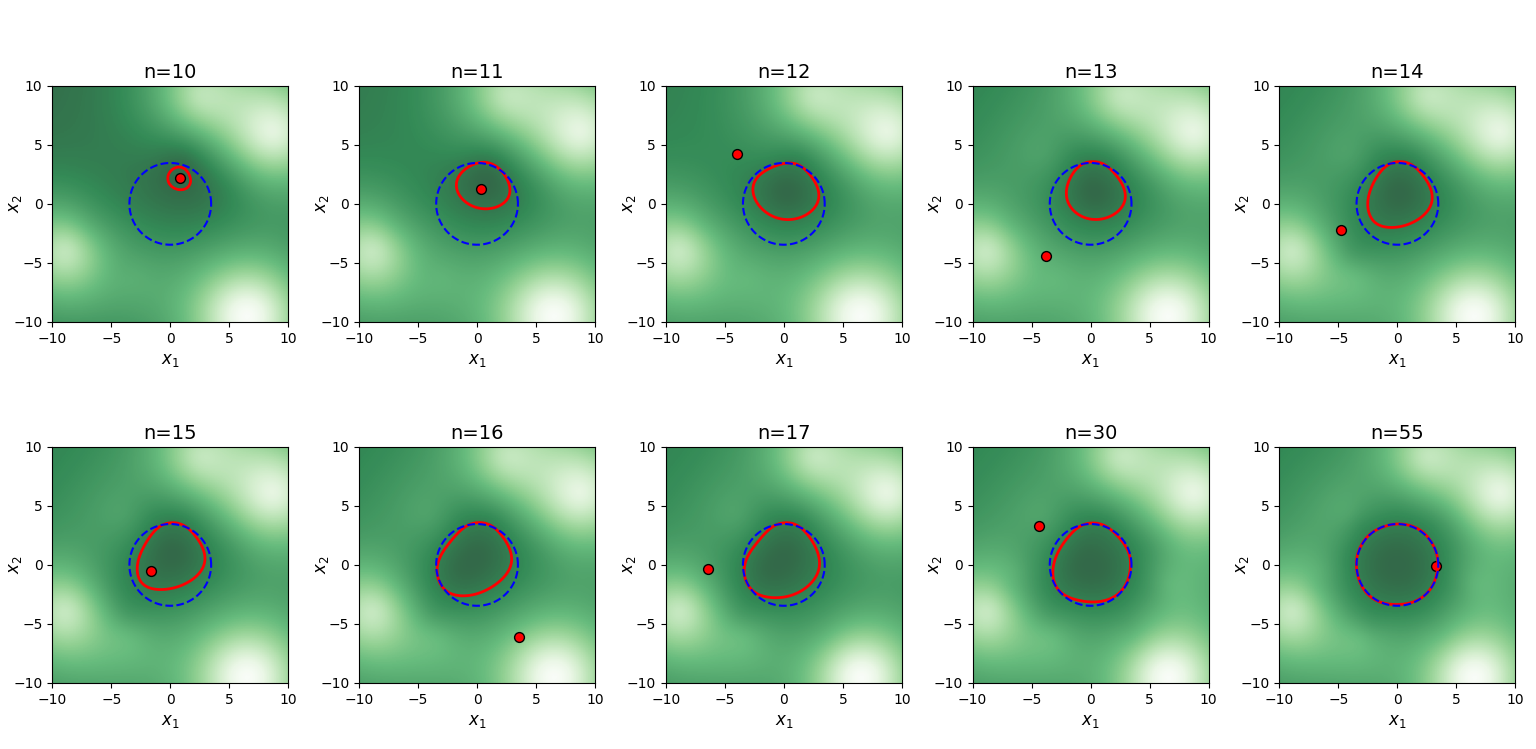}
	\caption{Predicted \( g(x_1, x_2) \): true failure boundary (blue dashed), approximated failure boundary (red solid), and optimal design points (red dots).}
	\label{fig:FBx}
\end{figure}

In this study, ten initial points are generated within the predefined domain using Latin Hypercube Sampling (LHS). The experimental design framework, as outlined in Algorithm~\ref{alg:design_optimization}, is employed to iteratively identify an optimal experimental design point in each cycle. This point is subsequently added to the initial point set for the next iteration. The computational metric for each iteration is the failure probability corresponding to the current surrogate model. After forty-five iterations, the failure probability essentially converges. The evolution of the failure boundaries and the corresponding model predictions from ten selected iterations are illustrated in Figure~\ref{fig:FBx}. It can be observed that the failure boundary gradually aligns with the true boundary, starting from a significant deviation initially. Notably, the experimental design points tend to concentrate in regions where the failure boundary is poorly fitted, demonstrating an ``active learning'' effect.

\begin{figure}[H]
  \centering
  \begin{minipage}{0.48\linewidth}
    \centering
    \includegraphics[width=1\linewidth]{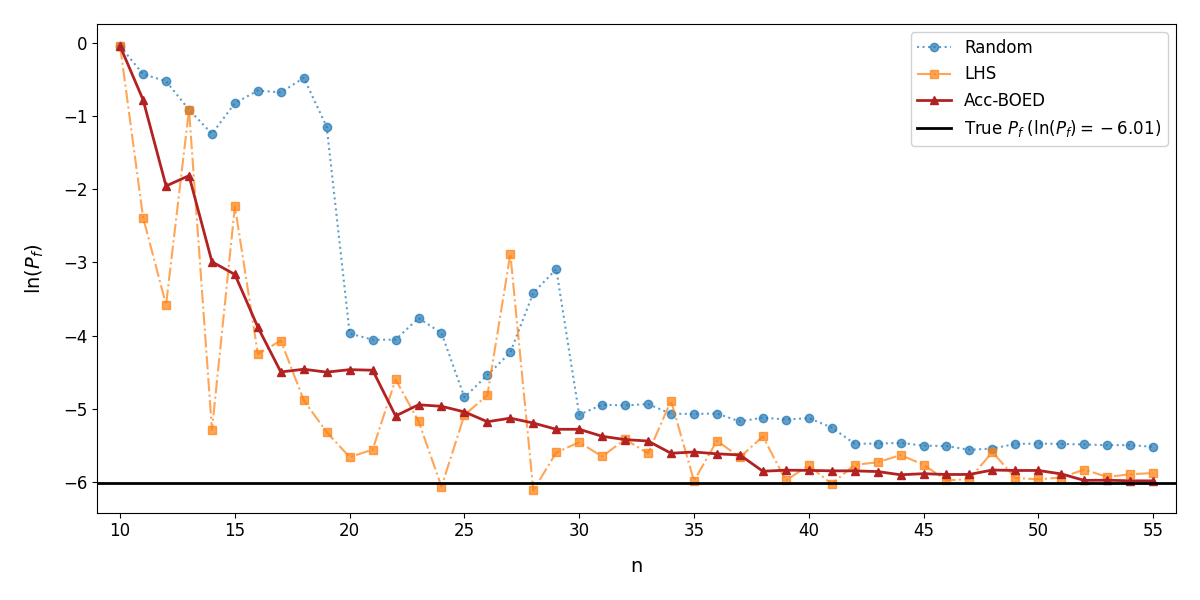}
    \label{fig:fb1}
  \end{minipage}
  \hfill
  \begin{minipage}{0.48\linewidth}
    \centering
    \includegraphics[width=1\linewidth]{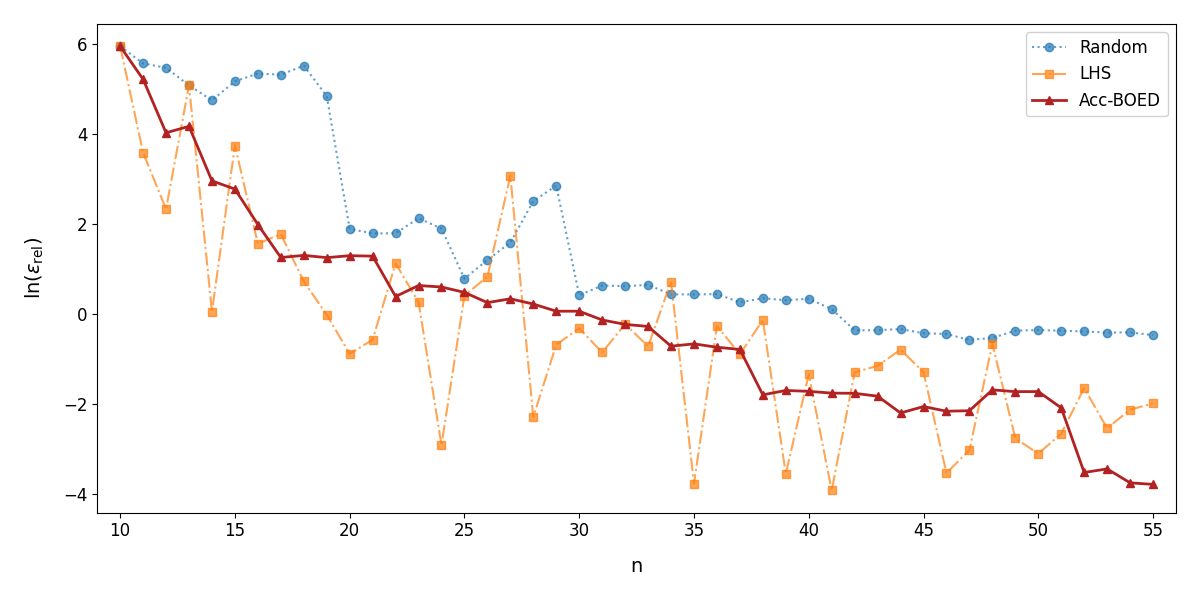}
    \label{fig:re1}
  \end{minipage}
  \caption{Logarithmic Evolution of Failure Probability (left) and Relative Error (right) for Three Methods}
  \label{fig:fb_re}
\end{figure}

The failure probability metric is computed using the Monte Carlo method. Since \( x_1 \) and \( x_2 \) are independently and identically distributed according to \( N(0, 1) \), a total of \( 10^6 \) samples are generated to calculate the number of failure points under different models. The true failure probability corresponding to the real model is \( 0.002460 \). The initial failure probability of the model, based on 10 data points, is \( 0.954054 \). After 46 iterations, the failure probability of the Acc-BOED method converges to the true failure probability. The evolution of the failure probabilities and relative errors for the Acc-BOED method, compared with the LHS method and random sampling, is illustrated in Figure~\ref{fig:fb_re}. For better visualization, both the failure probabilities and relative errors are logarithmically transformed. As observed in Figure~\ref{fig:fb_re}, the failure probability of the Acc-BOED method converges rapidly to the vicinity of the true failure probability and stabilizes, while its relative error steadily decreases, demonstrating the robustness of the method. In contrast to the inefficiency of random sampling and the instability of the LHS method, the Acc-BOED method exhibits significant advantages. The final failure probability at 55 points is \( 0.002516 \), which is very close to the true failure probability. The failure probabilities and relative errors for the three methods at 55 points are summarized in Table~\ref{tab:circle fp_re_table}. It can be seen that the failure probability of the Acc-BOED method is closer to the true failure probability, with a relative error of only \( 2.2784\% \), indicating that it is essentially consistent with the true failure probability.

\begin{table}[H]
\begin{center}
\caption{Failure Probability Estimation and Relative Error Comparison Across Different Methods for the Circle Problem with 55 Points}
\label{tab:circle fp_re_table} 
\begin{tabular}{cccc}
\toprule
& Failure Probability & Relative Error & \\
\midrule 
Ground Truth & 0.002460 & & \\
Acc-BOED & 0.002516 & 2.2764$\%$ & \\
Random scheme & 0.004006 & 62.8455$\%$ & \\
LHS scheme &  0.002799 & 13.7805$\%$ & \\
\bottomrule
\end{tabular}
\end{center}
\end{table}

\subsubsection{Four branch system}
Although the previous subsection demonstrated the feasibility of using the experimental design framework to address the failure probability estimation problem, as well as the effectiveness and superiority of the Acc-BOED method, the example presented was overly simplistic. The true limit state function was too straightforward, and the true failure boundary was excessively smooth, failing to adequately reflect the complexities inherent in failure probability estimation. In this subsection, we employ a widely-used benchmark model in failure probability analysis—the four-branch system—to evaluate our experimental design framework. The true state function of the system under this problem is defined as follows:
\begin{equation}
\label{eq four branch}
    g\left(x_1,x_2\right)=\min\begin{Bmatrix}3+0.1*(x_1-x_2)^2-(x_1+x_2)/\sqrt2\\3+0.1*(x_1-x_2)^2-(x_1+x_2)/\sqrt2\\(x_1-x_2)+7/\sqrt2\\(x_2-x_1)+7/\sqrt2\end{Bmatrix}
\end{equation}

Here, \( x_1 \) and \( x_2 \) are the inputs to the system, independently and identically distributed according to \( N(0, 1) \). The true failure boundary of the system is a four-branched shape with highly sharp corners, which is generally challenging to accurately simulate.

Given the complexity of the limit state function in this problem, we employ an experimental design framework to address it. Due to the challenges in simulating the failure boundary, a relatively large number of data points is required. We consider two cases: the first case involves 31 iterations with an initial dataset of 20 points, illustrating how the Acc-BOED method transitions from a poorly fitted failure boundary to the true failure boundary; the second case involves 61 iterations with an initial dataset of 50 points, demonstrating how the Acc-BOED accurately captures the complex sharp corners. The initial data points for both cases are generated using Latin Hypercube Sampling (LHS) within the two-dimensional domain \([-10, 10]\).

Similarly, the Monte Carlo Markov Chain (MCMC) method is employed to simulate the true failure probability. A total of \( 10^6 \) samples are randomly drawn from the \( N(0, 1) \) distribution, yielding a true failure probability of 0.00225. In both cases, we compare the evolution of failure probability estimates obtained from the Acc-BOED method, Latin Hypercube Sampling (LHS), and random sampling. 

\begin{figure}[H]
  \centering	\includegraphics[width=.8\linewidth]{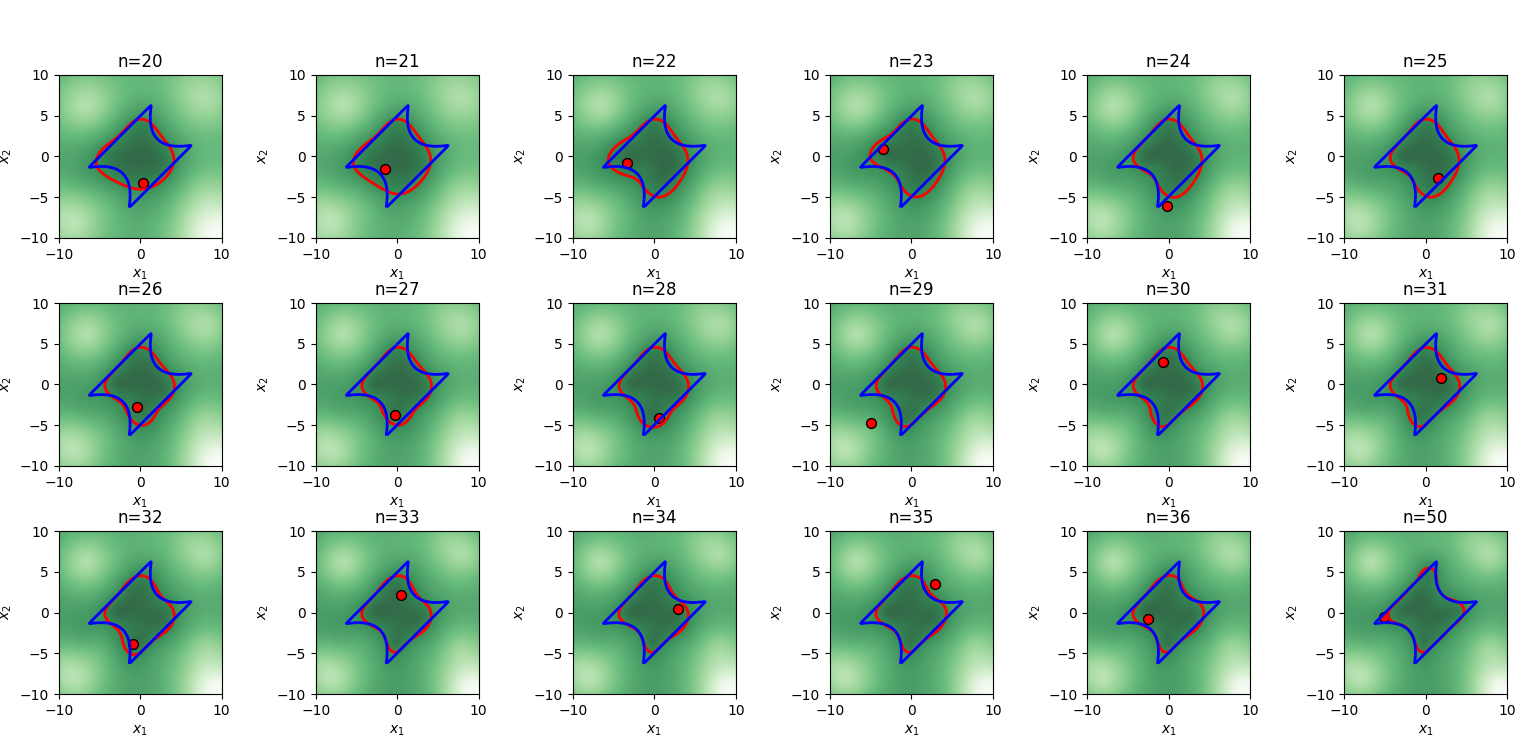}
	\caption{Predicted \( g(x_1, x_2) \)(from n=20): true failure boundary (blue dashed), approximated failure boundary (red solid), and optimal design points (red dots).} 
	\label{fig:fbs20-50}
\end{figure}

\begin{figure}[H]
  \centering
  \begin{minipage}{0.48\linewidth}
    \centering
    \includegraphics[width=1\linewidth]{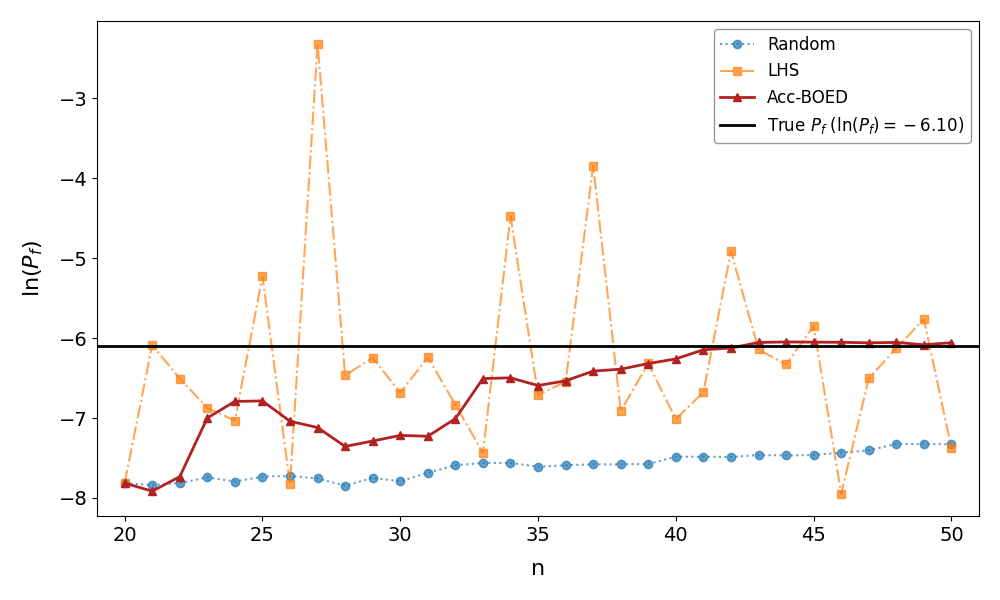}
    \label{fig:fb1}
  \end{minipage}
  \hfill
  \begin{minipage}{0.48\linewidth}
    \centering
    \includegraphics[width=1\linewidth]{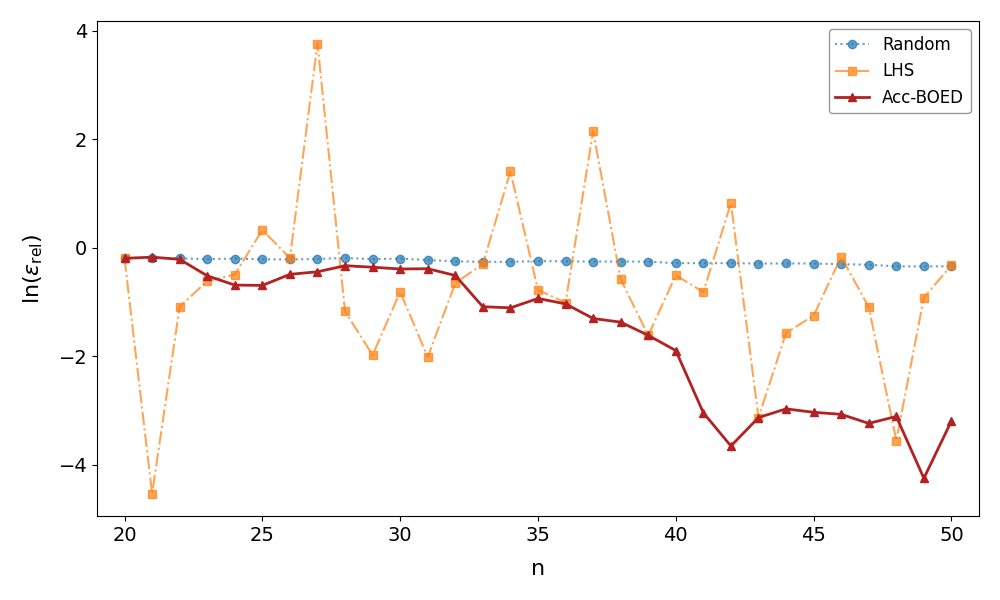}
    \label{fig:re1}
  \end{minipage}
  \caption{Logarithmic Evolution of Failure Probability (left) and Relative Error (right) for Three Methods(from n=20)}
  \label{fig:fb_re2}
\end{figure}

\begin{table}[H]
\begin{center}
\caption{Failure Probability Estimation and Relative Error Comparison Across Different Methods for the four branch system with 50 Points}
\begin{tabular}{cccc}
\toprule
& Failure Probability & Relative Errors & \\
\midrule 
Ground Truth & 0.00225 & & \\
Acc-BOED & 0.002342 & 4.0889$\%$ & \\
Random & 0.000658 & 70.7556$\%$ & \\
LHS &  0.000627 & 72.1333$\%$ & \\
\bottomrule
\label{tab:50four} 
\end{tabular}
\end{center}
\end{table}

The evolution of the failure boundary for the first case over 15 iterations is illustrated in Figure~\ref{fig:fbs20-50}. It can be observed that, with an initial dataset of 20 points, the approximated failure boundary (red solid line) significantly deviates from the true failure boundary (blue solid line). As the number of iterations increases, the experimental design points gradually capture information from poorly fitted regions, causing the approximated failure boundary to converge toward the true boundary. By the 50th data point, the approximated failure boundary almost completely aligns with the true failure boundary. The evolution of the logarithmic failure probabilities and relative errors over 31 iterations for the three methods is shown in Figure~\ref{fig:fb_re2}. This further demonstrates the clear advantage of the Acc-BOED method over LHS and random sampling. The failure probability of the Acc-BOED method converges more rapidly to the vicinity of the true failure probability, and its relative error consistently decreases, confirming the reliability of the Acc-BOED method. The final failure probabilities and relative errors for the three methods at 50 points are summarized in Table~\ref{tab:50four}. The failure probability of the Acc-BOED method (0.002342) is closer to the true failure probability (0.00225), with a relative error of only 4.0889\%, significantly outperforming LHS and random sampling.

\begin{figure}[H]
  \centering
	\includegraphics[width=.8\linewidth]{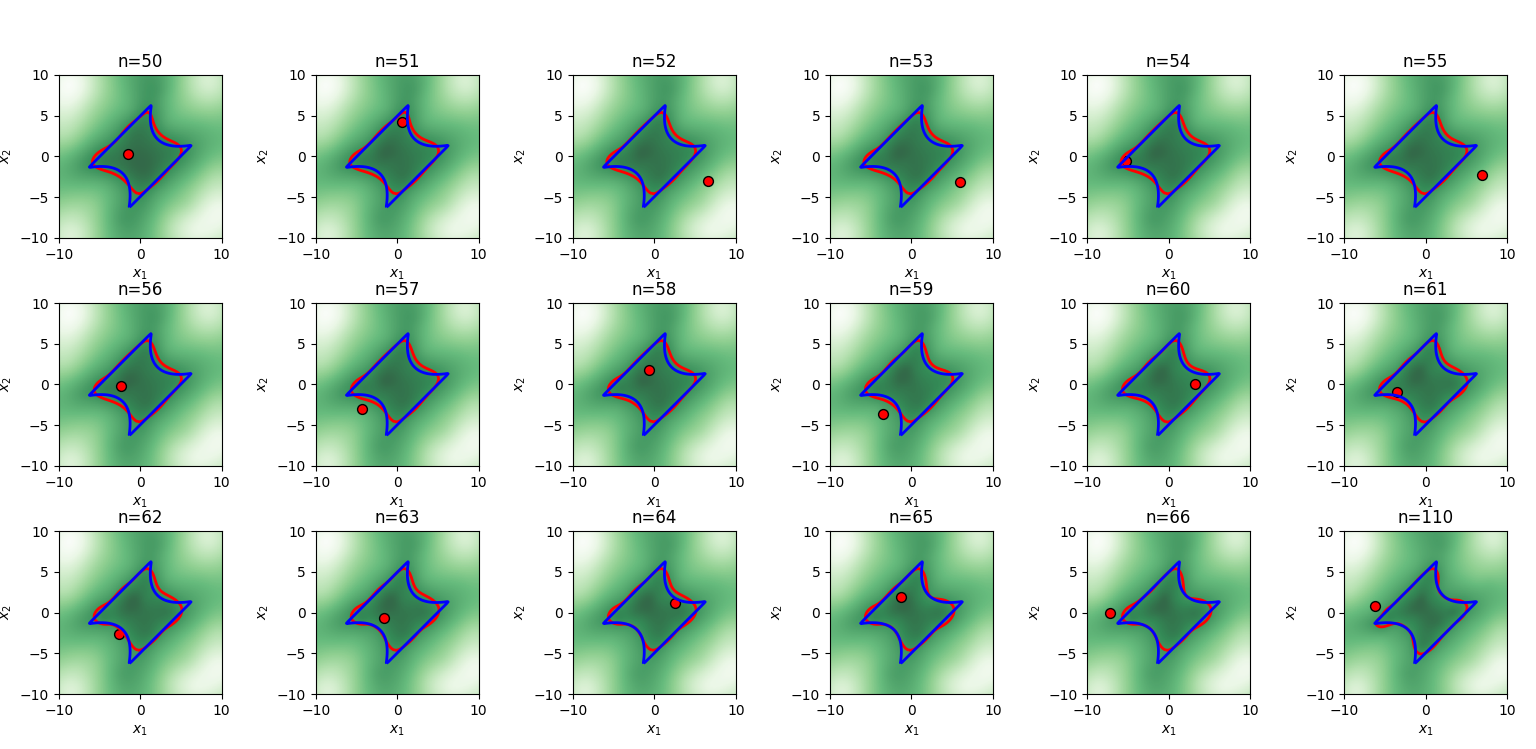}
	\caption{ Predicted \( g(x_1, x_2) \)(from n=50): true failure boundary (blue dashed), approximated failure boundary (red solid), and optimal design points (red dots).} 
	\label{fig:fb50-110}
\end{figure}

\begin{table}[H]
\begin{center}
\caption{Failure Probability Estimation and Relative Error Comparison Across Different Methods for the four branch system with 110 Points}
\begin{tabular}{cccc}
\toprule
& Failure Probability & Relative Errors & \\
\midrule
Ground Truth & 0.00225 & & \\
Acc-BOED & 0.002296 & 2.0444$\%$ & \\
Random & 0.001174 & 47.8222$\%$ & \\
LHS &  0.002351 & 4.4889$\%$ & \\
\bottomrule
\label{tab:110four}
\end{tabular}
\end{center}
\end{table}

\begin{figure}[H]
  \centering
	\includegraphics[width=.8\linewidth]{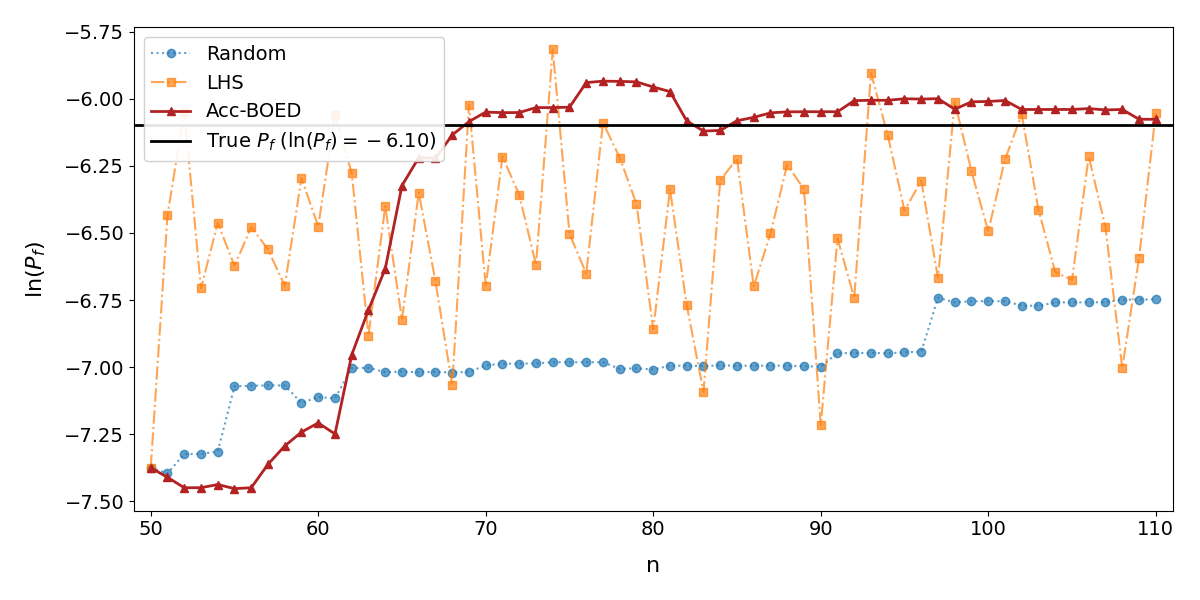}
	\caption{Evolution of failure probability estimations} 
	\label{fig:fp2}
\end{figure}

Similarly, the evolution of the failure boundary for the second case over 15 iterations is shown in Figure~\ref{fig:fb50-110}. It can be seen that, during the iterative process, the experimental design progressively captures information from the four corners, guiding the fitting of the failure boundary. Ultimately, the four corners are accurately fitted. The evolution of the failure probability over 60 iterations is illustrated in Figure~\ref{fig:fp2}, where the horizontal axis represents the size of the dataset, and the vertical axis represents the logarithmic failure probability. Compared to LHS and random sampling, the Acc-BOED method exhibits greater stability, with its failure probability curve gradually converging toward the true failure probability. In contrast, LHS shows excessive oscillations and instability, while random sampling is inefficient and struggles to converge to the true failure probability. The final failure probabilities and relative errors for the three methods at 110 points are summarized in Table~\ref{tab:110four}. At this stage, the relative error is reduced to 2.0444\%, and the failure probability is closer to the true value, indicating a better-fitted failure boundary and higher model accuracy. Although the numerical improvement is only approximately 2\%, the accuracy at the four corners is significantly enhanced. This highlights the complexity of the four-branch system, where the four corners are challenging to simulate and yield minimal improvements. Nevertheless, these results demonstrate the superiority and reliability of the Acc-BOED method.

\section{Conclusion}
\label{set:conclusion}
We propose an accelerated Bayesian optimal experimental design method based on conditional density estimation and informative data-set, named Acc-BOED. This method addresses the challenges of low simulation efficiency and expensive data acquisition by reformulating the original problems of efficient surrogate model construction, failure probability estimation, and parameter estimation as optimal experimental design problems within a Bayesian framework. 
By representing complex Gaussian random fields as the product of simpler Gaussian random fields and conditional density estimation models, the Acc-BOED method significantly reduces computational complexity and improves efficiency.
The Acc-BOED method optimizes the experimental design by maximizing the integral of the utility function, approximates the posterior distribution via conditional density estimation, and accelerates model fitting and integral calculations using covariance as a metric.
This approach provides a comprehensive BOED framework grounded in conditional density estimation, offering a robust solution for complex engineering problems.

Bayesian optimal experimental design methods based on conditional density estimation and related techniques are rapidly gaining attention in scientific and engineering fields due to their potential to accelerate simulations and improve efficiency. Beyond the conditional density estimation employed in this work, other advanced techniques such as normalizing flows and variational methods can be integrated to handle higher-dimensional functions and more complex uncertainty structures. Future work will explore these extensions to further enhance the applicability and performance of the Acc-BOED method.

\section*{Acknowledgement}

The work was supported by the Innovative Platform Project of the Department of Science and Technology of Hunan Province (2024JC1003).
Additional support was provided by the High Performance Computing Center of Central South University.
 
\bibliography{ref}

\begin{thebibliography}{10}

\bibitem{fisher1922mathematical}
Ronald~A Fisher.
\newblock On the mathematical foundations of theoretical statistics.
\newblock {\em Philosophical transactions of the Royal Society of London.
  Series A, containing papers of a mathematical or physical character},
  222(594-604):309--368, 1922.

\bibitem{atkinson1996usefulness}
Anthony~C Atkinson.
\newblock The usefulness of optimum experimental designs.
\newblock {\em Journal of the Royal Statistical Society Series B: Statistical
  Methodology}, 58(1):59--76, 1996.

\bibitem{pukelsheim2006optimal}
Friedrich Pukelsheim.
\newblock {\em Optimal design of experiments}.
\newblock SIAM, 2006.

\bibitem{fedorov2013theory}
Valerii~Vadimovich Fedorov.
\newblock {\em Theory of optimal experiments}.
\newblock Elsevier, 2013.

\bibitem{sim2005global}
Robert Sim and Nicholas Roy.
\newblock Global a-optimal robot exploration in slam.
\newblock In {\em Proceedings of the 2005 IEEE international conference on
  robotics and automation}, pages 661--666. IEEE, 2005.

\bibitem{de1995d}
P~Fernandes de~Aguiar, B~Bourguignon, MS~Khots, DL~Massart, and
  R~Phan-Than-Luu.
\newblock D-optimal designs.
\newblock {\em Chemometrics and intelligent laboratory systems},
  30(2):199--210, 1995.

\bibitem{lindley1968choice}
Dennis~V Lindley.
\newblock The choice of variables in multiple regression.
\newblock {\em Journal of the Royal Statistical Society: Series B
  (Methodological)}, 30(1):31--53, 1968.

\bibitem{lindley1972bayesian}
Dennis~Victor Lindley.
\newblock {\em Bayesian statistics: A review}.
\newblock SIAM, 1972.

\bibitem{chaloner1982optimal}
Kathryn~Mary Chaloner.
\newblock {\em Optimal Bayesian experimental design for linear models}.
\newblock Carnegie Mellon University, 1982.

\bibitem{pilz1991bayesian}
J{\"u}rgen Pilz.
\newblock Bayesian estimation and experimental design in linear regression
  models.
\newblock {\em (No Title)}, 1991.

\bibitem{clyde1995exploring}
Merlise~A Clyde, Peter M{\"u}ller, and Giovanni Parmigiani.
\newblock {\em Exploring expected utility surfaces by markov chains}.
\newblock Institute of Statistics \& Decision Sciences, Duke University, 1995.

\bibitem{bielza1999decision}
Concha Bielza, Peter M{\"u}ller, and David~R{\'\i}os Insua.
\newblock Decision analysis by augmented probability simulation.
\newblock {\em Management Science}, 45(7):995--1007, 1999.

\bibitem{stroud2001optimal}
Jonathan~R Stroud, Peter M{\"u}ller, and Gary~L Rosner.
\newblock Optimal sampling times in population pharmacokinetic studies.
\newblock {\em Journal of the Royal Statistical Society Series C: Applied
  Statistics}, 50(3):345--359, 2001.

\bibitem{muller2005simulation}
Peter M{\"u}ller.
\newblock Simulation based optimal design.
\newblock {\em Handbook of Statistics}, 25:509--518, 2005.

\bibitem{amzal2006bayesian}
Billy Amzal, Fr{\'e}d{\'e}ric~Y Bois, Eric Parent, and Christian~P Robert.
\newblock Bayesian-optimal design via interacting particle systems.
\newblock {\em Journal of the American Statistical association},
  101(474):773--785, 2006.

\bibitem{muller2006bayesian}
Peter M{\"u}ller, Don~A Berry, Andrew~P Grieve, and Michael Krams.
\newblock A bayesian decision-theoretic dose-finding trial.
\newblock {\em Decision analysis}, 3(4):197--207, 2006.

\bibitem{cook2008optimal}
Alex~R Cook, Gavin~J Gibson, and Christopher~A Gilligan.
\newblock Optimal observation times in experimental epidemic processes.
\newblock {\em Biometrics}, 64(3):860--868, 2008.

\bibitem{cavagnaro2010adaptive}
Daniel~R Cavagnaro, Jay~I Myung, Mark~A Pitt, and Janne~V Kujala.
\newblock Adaptive design optimization: A mutual information-based approach to
  model discrimination in cognitive science.
\newblock {\em Neural computation}, 22(4):887--905, 2010.

\bibitem{ryan2015fully}
Elizabeth~G Ryan, Christopher~C Drovandi, and Anthony~N Pettitt.
\newblock Fully bayesian experimental design for pharmacokinetic studies.
\newblock {\em Entropy}, 17(3):1063--1089, 2015.

\bibitem{toman1994efficiency}
Blaza Toman and Joseph~L Gastwirth.
\newblock Efficiency robust experimental design and estimation using a
  data-based prior.
\newblock {\em Statistica Sinica}, pages 603--615, 1994.

\bibitem{kadane2011bayesian}
Joseph~B Kadane.
\newblock {\em Bayesian methods and ethics in a clinical trial design}.
\newblock John Wiley \& Sons, 2011.

\bibitem{etzioni1993optimal}
Ruth Etzioni and Joseph~B Kadane.
\newblock Optimal experimental design for another's analysis.
\newblock {\em Journal of the American Statistical Association},
  88(424):1404--1411, 1993.

\bibitem{han2004bayesian}
Cong Han and Kathryn Chaloner.
\newblock Bayesian experimental design for nonlinear mixed-effects models with
  application to hiv dynamics.
\newblock {\em Biometrics}, 60(1):25--33, 2004.

\bibitem{wu2023large}
Keyi Wu, Thomas O’Leary-Roseberry, Peng Chen, and Omar Ghattas.
\newblock Large-scale bayesian optimal experimental design with
  derivative-informed projected neural network.
\newblock {\em Journal of Scientific Computing}, 95(1):30, 2023.

\bibitem{aushev2023online}
Alexander Aushev, Aini Putkonen, Gr{\'e}goire Clart{\'e}, Suyog Chandramouli,
  Luigi Acerbi, Samuel Kaski, and Andrew Howes.
\newblock Online simulator-based experimental design for cognitive model
  selection.
\newblock {\em Computational Brain \& Behavior}, 6(4):719--737, 2023.

\bibitem{valentin2024designing}
Simon Valentin, Steven Kleinegesse, Neil~R Bramley, Peggy Seri{\`e}s, Michael~U
  Gutmann, and Christopher~G Lucas.
\newblock Designing optimal behavioral experiments using machine learning.
\newblock {\em Elife}, 13:e86224, 2024.

\bibitem{orozco2024probabilistic}
Rafael Orozco, Felix~J Herrmann, and Peng Chen.
\newblock Probabilistic bayesian optimal experimental design using conditional
  normalizing flows.
\newblock {\em arXiv preprint arXiv:2402.18337}, 2024.

\bibitem{foster2019variational}
Adam Foster, Martin Jankowiak, Elias Bingham, Paul Horsfall, Yee~Whye Teh,
  Thomas Rainforth, and Noah Goodman.
\newblock Variational bayesian optimal experimental design.
\newblock {\em Advances in Neural Information Processing Systems}, 32, 2019.

\bibitem{dong2025variational}
Jiayuan Dong, Christian Jacobsen, Mehdi Khalloufi, Maryam Akram, Wanjiao Liu,
  Karthik Duraisamy, and Xun Huan.
\newblock Variational bayesian optimal experimental design with normalizing
  flows.
\newblock {\em Computer Methods in Applied Mechanics and Engineering},
  433:117457, 2025.

\bibitem{huan2013simulation}
Xun Huan and Youssef~M Marzouk.
\newblock Simulation-based optimal bayesian experimental design for nonlinear
  systems.
\newblock {\em Journal of Computational Physics}, 232(1):288--317, 2013.

\bibitem{wang2021explicit}
Hongqiao Wang and Xiang Zhou.
\newblock Explicit estimation of derivatives from data and differential
  equations by gaussian process regression.
\newblock {\em International Journal for Uncertainty Quantification}, 11(4),
  2021.

\bibitem{ambrogioni2017kernel}
Luca Ambrogioni, Umut G{\"u}{\c{c}}l{\"u}, Marcel~AJ van Gerven, and Eric
  Maris.
\newblock The kernel mixture network: A nonparametric method for conditional
  density estimation of continuous random variables.
\newblock {\em arXiv preprint arXiv:1705.07111}, 2017.

\bibitem{li2019computing}
Qianxiao Li, Bo~Lin, and Weiqing Ren.
\newblock Computing committor functions for the study of rare events using deep
  learning.
\newblock {\em The Journal of Chemical Physics}, 151(5), 2019.

\bibitem{ren2005transition}
Weiqing Ren, Eric Vanden-Eijnden, Paul Maragakis, et~al.
\newblock Transition pathways in complex systems: Application of the
  finite-temperature string method to the alanine dipeptide.
\newblock {\em The Journal of chemical physics}, 123(13), 2005.

\bibitem{phillips2005scalable}
James~C Phillips, Rosemary Braun, Wei Wang, James Gumbart, Emad Tajkhorshid,
  Elizabeth Villa, Christophe Chipot, Robert~D Skeel, Laxmikant Kale, and Klaus
  Schulten.
\newblock Scalable molecular dynamics with namd.
\newblock {\em Journal of computational chemistry}, 26(16):1781--1802, 2005.

\end{thebibliography}
\end{document}